\documentclass[acmlarge]{acmart}

\AtBeginDocument{%
  \providecommand\BibTeX{{%
    \normalfont B\kern-0.5em{\scshape i\kern-0.25em b}\kern-0.8em\TeX}}}

\setcopyright{none}

\usepackage{longtable}
\usepackage{graphicx}
\usepackage{subfigure}
\usepackage{array,tabularx,booktabs}
\usepackage{amsmath}
\usepackage{hyperref}
\newcolumntype{s}{>{\hsize=.5\hsize}X}

\usepackage{makecell}

\usepackage{multirow}
\usepackage{xcolor}[table,xcdraw]

\usepackage{color}
\definecolor{Orange}{rgb}{0.9,0.5,0}
\definecolor{NavyBlue}{rgb}{0.1, 0.4, 0.8}
\definecolor{Magenta}{rgb}{0.8, 0.1, 0.6}
\definecolor{Yellow}{rgb}{0.9, 0.8, 0.4}

\begin{document}

\title{A Survey on Deep Learning for Human Mobility}

\author{Massimiliano Luca}
\affiliation{%
  \institution{Fondazione Bruno Kessler (FBK)}
  \streetaddress{Via Sommarive, 19}
  \city{Povo - Trento}
  \country{Italy}}
\affiliation{%
  \institution{Free University of Bolzano}
  \streetaddress{Piazza Domenicani, 3}
  \city{Bolzano}
  \country{Italy}}
\email{mluca@fbk.eu}
\orcid{0000-0001-6964-9877}

\author{Gianni Barlacchi}
\authornote{Work done before joining Amazon.}
\affiliation{%
  \institution{Amazon Alexa}
  \streetaddress{}
  \city{Berlin}
  \country{Germany}}
\email{gianni.barlacchi@gmail.com}
\orcid{}

\author{Bruno Lepri}
\affiliation{%
  \institution{Fondazione Bruno Kessler (FBK)}
  \streetaddress{Via Sommarive, 19}
  \city{Povo - Trento}
  \country{Italy}}
\email{}
\orcid{}

\author{Luca Pappalardo}
\affiliation{
  \institution{Institute of Information Science and Technologies, National Research Council (ISTI-CNR)}
  \streetaddress{Via G. Moruzzi 1, 56124}
  \city{Pisa}
  \country{Italy}}
\email{luca.pappalardo@isti.cnr.it}
\orcid{0000-0002-1547-6007}

\renewcommand{\shortauthors}{Luca et al.}

\begin{abstract}

The study of human mobility is crucial due to its impact on several aspects of our society, such as disease spreading, urban planning, well-being, pollution, and more. 
The proliferation of digital mobility data, such as phone records, GPS traces, and social media posts, combined with the predictive power of artificial intelligence, triggered the application of deep learning to human mobility. 
Existing surveys focus on single tasks, data sources, mechanistic or traditional machine learning approaches, while a comprehensive description of deep learning solutions is missing.
This survey provides a taxonomy of mobility tasks, a discussion on the challenges related to each task and how deep learning may overcome the limitations of traditional models, a description of the most relevant solutions to the mobility tasks described above and the relevant challenges for the future.
Our survey is a guide to the leading deep learning solutions to next-location prediction, crowd flow prediction, trajectory generation, and flow generation.
At the same time, it helps deep learning scientists and practitioners understand the fundamental concepts and the open challenges of the study of human mobility.

\end{abstract}



\maketitle

\newpage
\section{Introduction}

Urban population is increasing strikingly and human mobility is becoming more complex and bulky, affecting crucial aspects of people lives such as the spreading of viral diseases (e.g., the COVID-19 pandemic) \cite{pepe2020covid, lai2019measuring, Ruktanonchai1465, kraemer2020effect, oliver2020mobile, pappalardo2020dataset}, the behavior of people in case of natural disasters \cite{10.1145/2970819, jiang2018deepurbanmomentum, wang2018coupling}, the public and private transportation and the resulting traffic volumes \cite{rossi2019modelling, ferretti2018weak, chen2019stay, khaidem2020optimize}, the well-being of citizens \cite{pappalardo2016analytical, voukelatou2020measuring, soto2011prediction}, the severity of air pollution, energy and water consumption \cite{bohm2021quantifying, stewart2018electric, nyhan2019quantifying}. 
Furthermore, crowds' movement between cities is influenced by migrations from rural to urban areas, such as those induced by natural disasters, climate change, and conflicts \cite{prieto2018gravity, ahmed2016multi, gray2012natural, reuveny2007climate, simini2020deep, luca2021leveraging, sirbu2020human}.

\begin{figure}[!htb]
	\centering
    \includegraphics[width=0.7\textwidth]{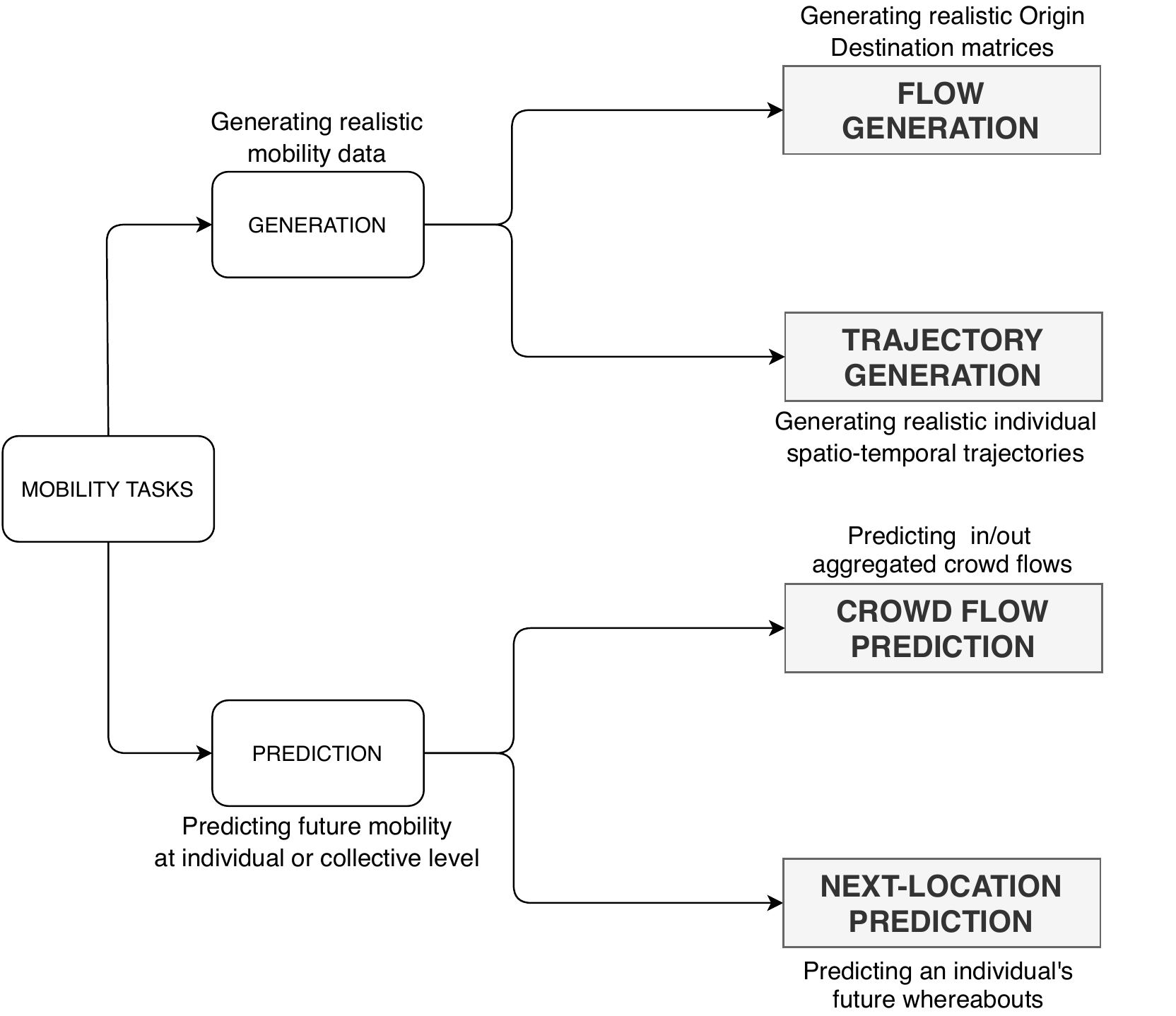}
\caption{A taxonomy of the mobility tasks we discuss in this survey. 
We classify mobility tasks in predictive, aiming at forecasting future mobility at an individual or collective level (Section \ref{sec:predictive_models}), and generative, aiming at generating realistic trajectories of mobility flows (Section \ref{sec:generative_models}). 
Among the predictive tasks, we cover \emph{i)} next-location prediction, the problem of forecasting future whereabouts given the mobility history of individuals (Section \ref{sec:next}), and \emph{ii)} crowd flow prediction, whose goal is to forecasting future aggregated flows given historical observations (Section \ref{sec:crowd}). 
On the other hand, we have two generative tasks: \emph{i)} trajectory generation aims at generating realistic individual trajectories (Section \ref{sec:task_generative}),  and \emph{ii)} flow generation whose goal is to generate realistic flows among locations on a geographic region (Section \ref{sec:flow_generative}). 
We use this taxonomy to map relevant works to the task they solve and shape this survey's structure.}
	\label{fig:taxonomy}
\end{figure}

Fortunately, policymakers are not unarmed in facing these challenges. 
The rise of ubiquitous computing (e.g., mobile phone, the Internet of Things, social media platforms) provides an always up-to-date and precise way to sense human movements at various temporal and spatial scales. 
Examples of mobility data include tracks from GPS devices embedded in smartphones \cite{zheng2010geolife, roma-taxi-20140717, moreira2013predicting, laurila2012mobile, epfl-mobility-20090224}, vehicles \cite{bazzani2010statistical, pappalardo2013understanding, gallotti2013entropic, pappalardo2015returners} or boats \cite{chang2010vessel, fernandez2017maritime, riveiro2018maritime, yan2020exploring}; records produced by the communication between phones and the cellular network \cite{blondel2015survey, gonzalez2008understanding}; and geotagged posts from social media platforms \cite{jurdak2015understanding, rebelo2015twitterjam, blanford2015geolocated, liao2019individual, cui2018social}. 
This deluge of digital data fostered a vast scientific production on various aspects of human mobility, such as the mining of trajectory data \cite{zheng2015trajectory, zhao2016urban, mazimpaka2016trajectory, wang2020big, jiang2013review}, the uncovering of the statistical patterns \cite{gonzalez2008understanding, song2010limits, blondel2015survey, barbosa2018human}, and the estimation of the privacy risk \cite{pellungrini2017data, demontoye2018privacy, rossi2015privacy, pellungrini2020modeling, perentis2017anonymous, fiore2019privacy}.
{\color{black}The development of powerful Artificial Intelligence (AI) techniques and the availability of big mobility data offered unprecedented opportunities for researchers to use Deep Learning (DL) approaches to solve mobility-related challenges. 
In this survey, we focus on DL solutions to predict or generate human movements and exclude other approaches solving other problems, such as semantic enrichment of mobility data (e.g., predicting the purpose of movement) \cite{rinzivillo2014purpose}, home location detection \cite{pappalardo2021evaluation}, and population inference \cite{deville2014dynamic}. 
In particular, we focus on two categories of tasks: predictive and generative (see Figure \ref{fig:taxonomy}).
We discuss two predictive tasks, namely next-location prediction and crowd flow prediction, and two generative tasks, namely trajectory generation and flow generation. 

} 

Next-location prediction is about forecasting which location an individual will visit given historical data about their mobility. 
It is crucial in many applications such as travel recommendation, location-aware advertisements and geomarketing, early warning of potential public emergencies, and recommendation of friends in social network platforms \cite{zhu2015modeling, burbey2012survey, wu2018location, zheng2018survey, zhao2020event}. 
Crowd flow prediction, instead, is the task of forecasting the incoming and outgoing flows of people on a geographic region, which has an impact on public safety, the definition of on-demand services, the management of land use, and traffic optimization \cite{jayarajah2018understanding, yin2020comprehensive, xie2020urban, ebrahimpour2019comparison, shi2019survey}.
Concerning generative tasks, trajectory generation deals with generating synthetic trajectories that can reproduce, realistically, the individual statistical patterns of human mobility \cite{barbosa2018human, hess2015data, solmaz2019survey, wang2019urban}.
{\color{black} Flow generation deals with generating realistic flows among locations, given their characteristics and the distance among them, and without any knowledge about the real flows.
}
{\color{black} Although approaches based on Machine Learning (ML) achieve good results in solving these four tasks \cite{burbey2012survey, wu2018location, zheng2018survey, yin2020comprehensive, xie2020urban, ebrahimpour2019comparison, shi2019survey}, multiple reasons pushed researchers to adopt DL techniques, such as the ability to automatically extract relevant patterns from (un)structured and heterogeneous data, and the outstanding results obtained in other fields (e.g., computer vision, natural language processing).}

{ \color{black} Several survey papers that provide interesting perspectives on single mobility tasks \cite{koolwal2020comprehensive, xie2020urban, ebrahimpour2019comparison, shi2019survey, barbosa2018human, hess2015data, solmaz2019survey, shin2020user}, mobility data sources \cite{xu2020survey, zheng2018survey}, or traditional ML approaches \cite{hess2015data, wang2019urban, toch2019analyzing}. 
A few surveys discuss DL approaches to spatio-temporal data mining \cite{wang2020deep}, tasks related to the smart city ecosystem \cite{chen2019survey} or traffic-related issues \cite{wang2019enhancing, jiang2021graph}, covering some aspects related to human mobility but without a specific focus on the challenges and solutions to mobility-related tasks. 
}

{ \color{black} In this survey, we discuss DL solutions to next-location prediction, crowd flow prediction, trajectory generation and flow generation, organizing them into a proper taxonomy and discussing why those solutions may overcome limitations of existing traditional models. 
In order to find relevant papers for these tasks, we searched on Scopus the following keywords: "crowd flow", "next-location", "flow prediction", "flow generation", "trajectory generation", "mobility generation", and "mobility prediction". 
Among the obtained results, we selected those using DL, which are commonly used in subsequent papers as baselines, which are seminal works for the task they address, and which add some novelty in terms of DL pipelines, how DL modules are combined and data handled.
For each task, we also describe the open mobility datasets used by the papers and the evaluation metrics typically adopted. 
In the Appendix, we provide a more detailed description of deep learning modules (Appendix \ref{app:dl}), datasets (Appendix \ref{app:data}), evaluation metrics (Appendix \ref{app:evaluation_metrics}),  and mobility patterns' (Appendix \ref{app:mobility_patterns}).}
In summary, this survey provides the reader with:
\begin{itemize}
\color{black} 
\item An introduction to the fundamental concepts and nomenclature of human mobility (Section \ref{sec:trajectory}) and the key ideas behind the DL modules (Sections \ref{sec:deep_learning_modules}). 

\item A taxonomy of tasks related to predicting and generating human movements, with a comprehensive discussion on next-location prediction (Section \ref{sec:next}), crowd flow prediction (Section \ref{sec:crowd}), trajectory generation (Section \ref{sec:task_generative}), and flow generation (Section \ref{sec:flow_generative}). 
For each task, we define the problem, discuss the DL modules used in the literature to address it, we highlight the advantages of using DL over traditional models and list the public datasets and evaluation metrics commonly used for each task. 

\item A discussion of the most interesting open challenges about the four tasks (Section \ref{sec:conclusion}). We also provide a GitHub repository (\href{https://bit.ly/DL4HM}{bit.ly/DL4HM}) where researchers can co-operate to update the list of relevant mobility datasets and papers.

\item An Appendix in which we discuss the characteristics of DL modules, the data sources, the public datasets, the evaluation metrics used in the selected papers, and the mobility patterns used to assess the realism of generative models.
\end{itemize} 

{ \color{black} The survey is structured as follows. In Section \ref{sec:background}, we provide the definition of spatio-temporal trajectories and spatial aggregations (Section \ref{sec:trajectory}), and we briefly discuss the key DL modules used by the models tackling human mobility challenges (Section \ref{sec:deep_learning_modules}). A detailed discussion of such modules and a more detailed overview of human-mobility well-known laws and patterns are discussed in Appendix \ref{app:mobility_patterns}. 

In Sections \ref{sec:predictive_models} and \ref{sec:generative_models}, we explore, respectively, the predictive and generative tasks summarized in Figure \ref{fig:taxonomy}. For each task, after discussing why it is a relevant problem, we formally define it. Then, after briefly presenting the traditional mechanistic and/or ML-based approaches, we describe in detail the aspects that are not captured or only partially captured by such models. Then, we highlight how DL-models overtake these limitations and which DL modules are commonly used to succeed. 
Finally, In Section \ref{sec:conclusion}, we discuss some of the open challenges and we derive some conclusions.}

\section{Background}
\label{sec:background}
{\color{black}
Here we introduce the notation used in the remainder of the paper.
In Section \ref{sec:trajectory}, we define key mobility concepts, and in Section \ref{sec:deep_learning_modules} we briefly introduce the DL concepts and notation used in this survey. 
}

\subsection{Spatio-temporal trajectories and spatial aggregations}
\label{sec:trajectory}

Mobility data describe the movements of a set of individuals during a period of observation. 
They are typically collected through electronic devices and stored in the form of spatio-temporal trajectories or mobility flows.

The trajectory of an individual is a sequence of records that allows for reconstructing their movements during the period of observation \cite{zheng2014urban, zheng2015trajectory}. 
Typically, each record contains the individual's identifier, a geographic location expressed as a spatial point, and a timestamp indicating when the individual went through that location. 

\begin{definition}
\label{def:trj_simple}
Let $u$ be an individual, a trajectory $T_u=\langle p_{1}, p_{2}, ..., p_{n_u} \rangle$ is a time-ordered sequence composed by $n_u$ spatio-temporal points visited by $u$.
A spatio-temporal point is a pair $p=(t, l)$, where $t$ indicates the time when point $l=(x, y)$ is visited by $u$, and $x$ and $y$ are spatial coordinates in a given reference system, e.g., latitude and longitude.
A semantic spatio-temporal point $p$ is a tuple $p = (o, t, l)$, where $t$ indicates the time when point $l=(x, y)$ is visited by $u$, $l$ is a pair of coordinates $(x,y)$, and $o$ is a parameter that brings some meaning to the point (e.g., home, workplace, or some other categories), if any.  
\end{definition}
{ \color{black} Typically, stay points (or stops) are detected on spatio-temporal trajectories to find locations in which users spend a minimum amount of time \cite{pappalardo2019scikit, zheng2015trajectory}. }
In some tasks, the geographic space is discretized by mapping the coordinates to a spatial tessellation, 
i.e., a covering of the bi-dimensional space using a countable number of geometric shapes called tiles, with no overlaps and no gaps.
For instance, for crowd flow prediction, a spatial tessellation is used to aggregate the flows of people moving among the tiles.

\begin{definition}
\label{def:tessellation}
Given an area $A$, a set of geographical polygons called \textit{tessellation}, $\mathcal{G}$,  is defined with the following properties: 
(1) $\mathcal{G}$ contains a finite number of polygons, $g_i$, called \textit{tiles}, $\mathcal{G} = \{g_i: i=1,...,n\} $; 
(2) the tiles are non-overlapping, $g_i \cap g_j = \emptyset, \, \forall i \neq j$;
(3) the union of all tiles completely covers $A$, $\bigcup_{i=1}^{n} g_i = A$. 
\end{definition}

The tiling of the geographic space aims at creating the covering of the entire area of interest using regular tiles, such as equilateral triangular, squared, quadrilateral, hexagonal tiles, or irregular tiles that define the shape of buildings, census cells, or administrative units.
A spatial join can then be used to associate each trajectory's point or stay point with the tile that contains it.
Since the tessellation has no overlapping tiles and no gaps, each point is assigned only to one tile.
Further details may be found in Appendix \ref{app:tessellations}.

\subsection{Deep Learning Modules}
\label{sec:deep_learning_modules}
{ \color{black}
Here we introduce the DL notation used in the remainder of the paper. For a detailed description of the DL modules introduced here, see Appendix \ref{app:dl}. }

\textbf{FCs:} Fully Connected networks (FCs) consist of a series of fully-connected layers, in which all the neurons are connected to those in the next layer.
FCs are universal approximators (i.e., can learn any representation function) \cite{goodfellow2016deep}. 
{\color{black} In human mobility tasks, FC networks are commonly used to capture the impact on individual or collective mobility of external features and/or preferences (e.g., weather conditions, presence of public events)}.

\textbf{RNNs, LSTMs and GRUs:} Recurrent Neural Networks (RNNs) \cite{10.5555/104279.104293} can efficiently deal with sequential data, {\color{black} and they are used to capture spatial and temporal patterns in mobility tasks.
For example, RNNs are used in next-location prediction to find periodic trajectories patterns, represented as stay point sequences.
In crowd flow prediction, RNNs are used to capture the flows' temporal patterns.
However, RNNs suffer from the vanishing gradient problem \cite{kolen2001gradient} and cannot propagate information found at early steps, losing relevant information at the beginning of a sequence when it is time to analyze its end \cite{kolen2001gradient}.  Long-short-term Memory networks (LSTMs) \cite{hochreiter1997long} and Gated Recurrent Units (GRUs) \cite{cho-etal-2014-learning} are two gate implementations that mitigate this problem}.
Appendix \ref{app:rnns} provides further details and references on RNNs, LSTMs and GRUs.

\textbf{Attention mechanisms:} These mechanisms are based on the idea that, when dealing with a large amount of information, our brains focus on the most significant parts and consider all the others as background information. {In attention mechanisms, \color{black} the information in input is scored according to the context (e.g., attention maps), and the model focuses more on the information with high scores.}
{\color{black} In human mobility, attention is widely used for next-location prediction and crowd flow prediction to capture user preferences and highlight relevant historical patterns, respectively. 
Further details can be found in Appendix \ref{app:attention}.}

{ \color{black}  \textbf{CNNs:} Similarly to the visual cortex \cite{hubel1959receptive, hubel1962receptive}, Convolutional Neural Networks (CNNs) are made of neurons that react only to certain stimuli in a restricted region of the visual field \cite{krizehvsky2012imagenet}.
They are effective in computer vision applications such as object recognition \cite{simonyan2014very, ren2015fast}, image classification and segmentation \cite{krizehvsky2012imagenet, farabet2012learning}, movement or event recognition \cite{toshev2014deeppose}, and more \cite{khan2020survey}.
In human mobility tasks, CNNs are widely used to capture spatial patterns in the data, especially in crowd flow prediction where the distribution of people on a geographic region is represented as an image. 
Additional information on CNNs can be found in Appendix \ref{app:cnns}.
}

{ \color{black} \textbf{Generative Models:} in human mobility, Variational AutoEncoders (VAEs) and Generative Adversarial Networks (GANs) are used to generate realistic trajectories (i.e., synthetic trajectories that realistically reproduce mobility patterns).
VAEs transform input data (e.g., trajectories) from a high-dimensional space to a low-dimensional space, encoding samples as a distribution \cite{kingma2013auto}.}
GANs \cite{goodfellow2014generative} 
set up a game between a generator (e.g., a neural network) and a discriminator (e.g., a classifier). 
The generator's goal is to generate realistic data to fool the discriminator, whose purpose is to classify real and fake data and provide feedback to the generator to improve the realism of the generated data. 
Details about VAEs and GANs can be found in Appendix \ref{app:generative}.

\section{Predictive Models}   
\label{sec:predictive_models}

{\color{black} 
The goal of predictive models for human mobility is to forecast future whereabouts, either at the individual or collective level. 
At the individual level, next-location predictors forecast an individual's future whereabouts, based on their historical observations (Section \ref{sec:next}).
At the collective level, crowd flow predictors forecast the amount of people moving from or to geographic locations given historical information about aggregated crowd flows (Section \ref{sec:crowd}). 
}

{\color{black} 
In this Section, we describe both tasks, discussing how DL brings significant improvements with respect to traditional approaches, and describing the relevant state-of-the-art solutions to each task, with a reference to the public datasets and the metrics used for training and testing the models.  
}

\subsection{Next-Location Prediction}
\label{sec:next}
Predicting individuals' future locations is relevant in multiple applications such as monitoring public health  \cite{barlacchi2017you, canzian2015trajectories}, well-being  \cite{pappalardo2016analytical, voukelatou2020measuring}, and traffic congestions \cite{shi2019survey}, and to improve travel recommendation,  geomarketing, and link prediction in social network platforms \cite{zhu2015modeling, burbey2012survey, wu2018location, zheng2018survey, zhao2020event}. 
Next-location predictors may help policymakers organize the public transportation network, urban planners decide a city's future developments, and transportation companies provide citizens with a better service in terms of traffic reduction and ease of mobility. {\color{black} Predicting an individual's next location is challenging because it requires capturing the spatial and temporal patterns that characterize human habits \cite{barbosa2018human}, and combining heterogeneous data sources to model multiple factors influencing human displacements (e.g., weather, transportation mode, presence of POIs).}

 \textbf{Problem Definition.}
 {\color{black}
Next-location prediction consists of forecasting the next location (stay point) an individual will visit in the future, given their historical mobility data.
Formally, let $u$ be a user, $T_u$ their trajectory, and  $p_{t} \in T_u$ $u$'s current location, next-location prediction aims at predicting $u$'s next destination $p_{t+1}$.
This problem is treated in two ways: \emph{(i)} as a multi-class classification task, in which we have as many classes as locations and we aim at predicting the next visited location $p_{t+1}$; or \emph{(ii)} as a regression task, predicting $p_{t+1} = (x_{t+1}, y_{t+1})$, where $x_{t+1}$ and $y_{t+1}$ are the next location's geographic coordinates. 
A variant of next-location prediction aims at forecasting the next Point Of Interest (POI) $p_{t+1}$ an individual $u$ will visit given their trajectory $T_u$. 
Regardless of the specific definition, next-location predictors output a ranking of the probability of each location to be $u$'s next destination. 
}

\textbf{DL vs Traditional approaches.} Next-location prediction has been widely explored prior to the DL explosion using probabilistic or pattern-based approaches, which can work with a reasonably small amount of data \cite{burbey2012survey, zheng2018survey}. 
In a seminal work, Calabrese et al. \cite{calabrese2010human} propose a probabilistic model combining people's trajectories and geographical features such as land use, POIs, and distance of trips. 
Ashbrook et al.  \cite{ashbrook2002learning} cluster GPS data into meaningful locations and incorporate them into a Markov model to predict individuals' future movements.
Gambs et al. \cite{gambs2010show} introduce a Mobility Markov Chain (MMC) in which states represent POIs and transitions between states correspond to a movement between two POIs \cite{gambs2010show, gambs2012next}.
Among the pattern-based approaches, Monreale et al. \cite{monreale2009wherenext} develop a trajectory pattern mining algorithm to represent movement patterns as sequences of regions frequently visited with a typical travel time. 
Although traditional approaches achieve good performance with a small amount of data, they have substantial limitations. 
Notably, they require a considerable effort in feature engineering and cannot capture long-range temporal and spatial dependencies \cite{sabarish2015survey}. 

\begin{figure}[!htb]
	\centering
    \includegraphics[width=1\textwidth]{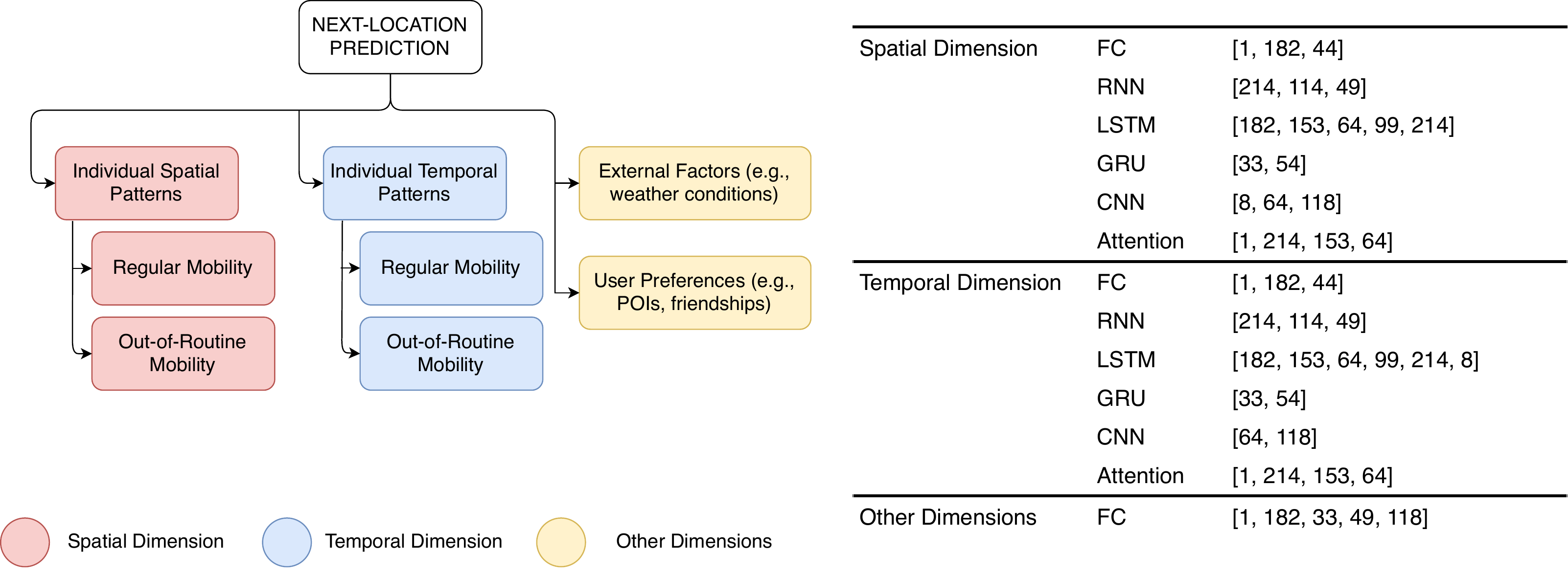}
\caption{\textbf{a)} The aspects next-location predictors should capture regarding the spatial dimension (red), the temporal dimension (blue), and the social and geographic dimensions (yellow) of human mobility data. 
\textbf{b)} DL modules that allow to capture each dimension, with the reference to the selected papers in the literature that implement these modules.}
	\label{fig:next_location_specialization}
\end{figure}

{ \color{black} 
Next-location predictors should capture the spatial, temporal and social-geographic dimensions of human mobility (Figure \ref{fig:next_location_specialization}). 
Regarding the spatial and temporal dimensions, predictors must capture at the same time spatial and temporal regularities hidden in human movements, as well as the tendency of people to get out of the routine. 
At the same time, predictors should capture the impact of external factors and individual preferences on the decision to move (e.g., weather conditions, preference for certain POIs, influence of friendships). 
Traditional approaches can only partially capture these aspects, and they particularly struggle in capturing complex sequential patterns in the data.
DL approaches overtake these issues by using mechanisms such as RNNs, LSTMs, GRUs, FCs, attention mechanisms, and CNNs to capture temporal, spatial, and social-geographic patterns in the data. 
}

{ \color{black} \textbf{Datasets and Evaluation Metrics.}
Next-location predictors are mainly trained and tested on public datasets of check-ins, generally coming from geosocial network platforms (e.g., Gowalla, Foursquare), and GPS traces, collected with smartphones or vehicles on-board GPS devices.
Examples of check-ins' datasets widely used for next-location prediction are Cho et al. \cite{cho2011friendship}, Feng et al. \cite{feng2018deepmove}, and Yang et al. \cite{yang2019revisiting}.

Examples of commonly used trajectory datasets are taxi traces collected in Porto, Portugal \cite{moreira2013predicting} and San Francisco, USA \cite{epfl-mobility-20090224}.
We refer to Appendix \ref{app:gps} and \ref{app:social_media} for details on the check-in and trajectory datasets mentioned above.

Whenever next-location prediction is intended as a regression task, predictors are evaluated using the Haversine distance or the equirectangular distance between the actual location and the predicted one (see Appendix \ref{app:distance_metrics} for details).
When next-location prediction is intended as a multi-class classification task, predictors are evaluated with accuracy (ACC/ACC@k), recall (Rec@k), F1-score (F1@k), Mean Average Percentage Error (MAPE) and/or Area Under the Curve (see Appendix \ref{app:classification_metrics} for details).
}

\indent {\bf DL approaches. }
{ \color{black}  
Table \ref{tab:nextloc} contains the selected DL approaches to next-location prediction, with the corresponding DL modules, datasets and evaluation metrics used, and link to an implementation (if available).}

\begin{table}[]
    \scriptsize
    \begin{tabularx}{\textwidth}{ c c c c c c c c X}
        \toprule
       & \scriptsize \textbf{Reference}   &   \scriptsize \textbf{Name} & \textbf{Year}       &   \scriptsize \textbf{DL Modules} &   \scriptsize \textbf{Evaluation}       &   \scriptsize \textbf{Dataset} &   \scriptsize \textbf{Code (https://bit.ly)} \\ \midrule 
        
         \multirow{14}{*}{\rotatebox[origin=c]{90}{\bf Next-Location Prediction}} & {\color{black} \emph{Abideen et al.}\cite{abideen2021deep}  } &  DWSTTN & 2021  & Encoder, Decoder, Attention, FC & Distance  & \cite{moreira2013predicting}  & - \\
         & {\color{black}\emph{Tang et al.}\cite{tang2021trip} }  & CLNN   & 2021  & LSTM, Embedding, FC  & Distance  & \cite{moreira2013predicting}   & - \\
        
         &{\color{black} \emph{Bao et al.}\cite{bao2020bilstm}}   & BiLSTM-CNN  &  2020 & Embedding, BiLSTM, CNN & ACC@k & -  \\
         
         & \emph{Chen et al.}\cite{chen2020context}   & DeepJMT  & 2020  & GRU, FC, Encoder  & ACC@k  & \cite{yang2019revisiting}  & - \\
         & \emph{Yang et al.}\cite{yang2020location}   & Flashback & 2020  & Attention, RNN & ACC@k & \cite{cho2011friendship} & \href{https://bit.ly/Flashback-1}{Flashback-1} \\
         & \emph{Ebel et al.}\cite{ebel2020destination}   & - & 2020  & RNN, FC, Embedding  & Distance & \cite{moreira2013predicting, epfl-mobility-20090224} & - \\

         & \emph{Rossi et al.}\cite{rossi2019modelling} & - & 2019 & Attention, LSTM & Distance & \cite{moreira2013predicting, tlc, epfl-mobility-20090224} & - \\
         & \emph{Gao et al.}\cite{gao2019predicting} & VANext  & 2019  & CNN, GRU, Attention & ACC@k & \cite{cho2011friendship} & - \\
         
         & \emph{Kong et al.}\cite{kong2018hst} & HST-LSTM & 2018 & LSTM & ACC & - &  \href{https://bit.ly/HST-LSTM}{HST-LSTM} \\

         & \emph{Lv et al.}\cite{lv2018t} & T-CONV & 2018 & CNN, FC &  Distance & \cite{moreira2013predicting} & \href{https://bit.ly/T-CONV}{T-CONV}\\
         
         & \emph{Feng et al.}\cite{feng2018deepmove} & DeepMove & 2018 & Attention, GRU, FC &  ACC & \cite{feng2018deepmove} & \href{https://bit.ly/DeepMove}{DeepMove}\\


         & \emph{Yao et al.}\cite{yao2017serm} & SERM & 2017 & LSTM & ACC@k & - & \href{https://bit.ly/SERM-Repo}{SERM-Repo} \\
         
         & \emph{Liu et al.}\cite{liu2016predicting} & ST-RNN  & 2016  & RNN & Rec@k, F1@k, MAPE, AUC & \cite{cho2011friendship, gtd} & \href{https://bit.ly/STRNN}{STRNN} \\
         
         & \emph{De Br{\'e}bisson et al.}\cite{de2015artificial} & -  & 2015  & FC &  Distance & \cite{moreira2013predicting} & \href{https://bit.ly/next-loc-1}{next-loc-1} \\
        
    \bottomrule    
     \end{tabularx} 
     \caption{List of the selected papers tackling next-location prediction. 
     For each paper, we describe the name of the corresponding proposed model (if any), the year of publication, the DL modules used in the proposed solution, the metrics used for performance evaluation, the link to the public datasets used for training and testing the model, and the link to a repository with the code (if any). 
     The papers are sorted by year of publication in decreasing order.}
     \label{tab:nextloc}
\end{table}
De Brébisson et al. \cite{de2015artificial} use an FC to predict taxi's passenger drop-off locations. The input data consist of trajectories represented as a variable-length sequence of GPS points and other meta-information, such as departure time, driver identity, and client information. 
The model performance is evaluated on the dataset of taxis in Porto \cite{moreira2013predicting} in terms of equirectangular distance to the actual visited location. However, it cannot take into consideration the temporal dimension of the mobility data.

ST-RNN (Spatial Temporal Recurrent Neural Networks) \cite{liu2016predicting} overtakes this issue by extending RNN with time- and spatial-specific transition matrices. 
Each RNN's layer learns an upper and lower bound for the temporal and spatial matrices through linear interpolation. 
The model is evaluated on the datasets of Gowalla \cite{cho2011friendship} and the Global Terrorism Dataset (GTD) \cite{gtd} using F1-score, Rec@k, MAPE and AUC. 

DeepMove \cite{feng2018deepmove} is an attentional recurrent network for mobility prediction from lengthy and sparse trajectories. 
First, historical and current trajectories are passed to a multi-modal embedding module to create a dense representation of the spatio-temporal and individual-specific features. Historical trajectories are handled by an attention mechanism to extract mobility patterns, while a GRU handles current trajectories. The output of the multi-modal embedding, the GRU, and the attention mechanism are concatenated and passed to an FC to predict an individual's next-location. 
DeepMove is evaluated, using ACC@k, on Foursquare data \cite{feng2018deepmove}, private mobile phone data, and a private dataset from a popular Chinese social network platform.

HST-LSTM (Hierarchical Spatial-Temporal LSTM) \cite{kong2018hst} aims at predicting an individual's short-term next-location. 
First, the authors design an ST-LSTM (Spatial Temporal LSTM), which combines the spatial and temporal characteristics of a trajectory using an LSTM. Then,
ST-LSTM is extended into HST-LSTM, which models periodic patterns using an encoder-decoder module. 
The encoder encodes the locations visited by a user in a given time span and area of interest, while 
the decoder predicts the possible areas of interest the user will visit next. 
HST-LSTM is evaluated using ACC on private data from Baidu.

T-Conv \cite{lv2018t} 
treats trajectories as images and handle them using CNNs to capture the spatial patterns at different scales.  
The output of the CNN, the trajectory's starting date-time, and other personal information about the user are passed to an FC that handles the prediction. 
T-Conv is evaluated using the datasets of taxis in Porto \cite{moreira2013predicting} and the Haversine distance.

Rossi et al. \cite{rossi2019modelling} propose an LSTM network equipped with a self-attention module to predict the coordinates of a taxi's next drop-off location.
Locations are enriched with geographical data to describe the surrounding area of the location semantically. 
The model is tested using the Haversine distance and on the datasets of taxis in Porto \cite{moreira2013predicting}, New York City \cite{tlc}, and San Francisco \cite{epfl-mobility-20090224}.

Another strand of research focuses on predicting the next POI an individual will visit using semantic trajectories.
For example, SERM (Semantics-Enriched Recurrent Model) \cite{yao2017serm} relies on an embedding layer to represent the timestamp, the location, and the keywords of a social media post concisely. 
Both the user's trajectory and the embedding are fed into an LSTM responsible for predicting the next POI. 
SERM is evaluated using ACC@k on Foursquare check-ins in New York City \cite{zhang2014splitter} and tweets in Los Angeles \cite{zhang2016gmove}.

In VANext (Variational Attention based Next Location) \cite{gao2019predicting}, the historical trajectories and the current one are embedded using two separate causal encoders to represent the semantic relationships among POIs. 
The encoded historical trajectories are passed to a CNN; the encoded current trajectory is passed to a GRU. 
The outputs of the CNN and the GRU are passed to an attention mechanism, which detects the most similar historical trajectory to the current one and passes it to an FC that predicts the individual's next POI. 
VANext is evaluated using ACC@k on the datasets of Gowalla \cite{cho2011friendship} and Foursquare for Singapore and New York City.

Flashback \cite{yang2020location} is based on an RNN and the concept of flashback, a technique that uses a sparse semantic trajectory to predict the next POI by looking for similar trajectories in terms of temporal characteristics. 
Flashback also uses an embedding to model the preferences of individuals to visit specific POIs. 
The outputs of the RNN and the embedding are passed to an FC that predicts the next visited POI. 
Flashback is evaluated using ACC@k on check-ins from Gowalla \cite{cho2011friendship} and Foursquare.

DeepJMT (Deep Model for Joint Mobility and Time) \cite{chen2020context} can predict an individual's next POI as well as when they will visit it. 
The model is based on four pipelines: a sequential dependency encoder, a spatial context encoder, a periodicity context extractor, and a social-temporal context extractor.
The sequential dependency encoder is a hierarchical GRU that takes as input an embedding of a user's trajectories. 
The high-level GRU captures the transitions between trajectories; while the low-level GRU models the transition within a trajectory. 
The spatial context extractor determines the dynamic influence of spatial neighbors, modeled as a graph in which nearest points influence the final prediction.
The periodicity context extractor is an attentional GRU aiming at extracting periodicity patterns from an individual's historical trajectories. 
The social-temporal context extractor leverages social relationships using an FC and pooling functions to facilitate both next-POI and time prediction.
Finally, the outputs of the four modules are concatenated to generate the prediction. 
DeepJMT is evaluated using ACC@k on Foursquare check-ins in New York City \cite{yang2019revisiting}, Tokyo \cite{yang2019revisiting} and Istanbul.


Ebel et al. \cite{ebel2020destination} propose a model to predict a vehicle's destinations and routes, given a partial trajectory and contextual data (e.g., day, time, weather). 
First, the area is tessellated and GPS points are assigned to these tiles using a k-d tree-based space partitioning method. 
The model is based on two main modules. 
The first module is an RNN that takes as input the mapped trajectory; while the second module is an FC that takes as input the embedded contextual data.
The two modules' outputs are merged and passed to an additional FC that produces the individual's probabilities to end a trip into a specific tile. 
The model is evaluated with the mean Haversine distance and the distance to the actual arrival point on taxis' traces in Porto \cite{moreira2013predicting} and San Francisco \cite{epfl-mobility-20090224}.

{\color{black}
BiLSTM-CNN \cite{bao2020bilstm} relies on a spatial clustering algorithm to derive areas of interest from POIs and uses bi-LSTMs and CNNs to predict the next area an individual will visit. 
The historical mobility data of people are passed into a bi-directional LSTM and then to a CNN to capture the overall spatial and temporal patterns. 
The CNN's output is then passed to an FC handling the prediction of the next location.
The model is evaluated with ACC@1 and ACC@5 on a private dataset of Weibo check-ins collected in Wuhan, China. 

CLNN (Classification Learning Neural Network) \cite{tang2021trip} relies on an information extraction module that extracts coordinates, date, time and driver characteristics, POIs and historical information from mobility data (e.g., similar trajectories). 
An LSTM is used to process the coordinates. 
Date, time, driver characteristics and POIs are embedded into a dense representation.
Two FCs process the embedded and historical information, respectively.
The outputs of the two FCs and the LSTM are fused together with a weighted sum, whose output is fed into an FC to predict the coordinates of the next taxi's destination. 
The model is evaluated using the mean Haversine distance on the dataset of taxis in Porto \cite{moreira2013predicting}.

DWSTTN (Deep Wide Spatio Temporal Transformer Network) \cite{abideen2021deep} is a transformer-like architecture that predicts a taxi's next destination using historical pick-up and drop-off observations and taxis' features and preferences. 
The model consists of two identical parts, the encoder and the decoder.
The encoder is responsible for the learning phase, the decoder for predicting the coordinates of the taxi's next location. 
In both, information about a taxi (e.g., identifier and stand identifier) and temporal information (e.g., weekday, hour, day type) are passed to an embedding layer (temporal transformer), while information about locations is embedded by a spatial transformer. 
The outputs of the temporal and spatial transformers are passed to two attention mechanisms to extract further relevant spatial and temporal information. 
The outputs of the attention mechanisms are fused together using a weighted sum, whose output is passed to an FC.
In the encoder, the FC outputs the learned features' representation; in the decoder, the FC outputs the next location's coordinates.
The model is evaluated using the Haversine distance on the datasets of taxis in Porto \cite{moreira2013predicting} and New York \cite{tlc}.
}

\subsection{Crowd Flow Prediction}
\label{sec:crowd}
Crowd flow prediction is the problem of forecasting the incoming and outgoing flows of locations in a geographic region, usually split into tiles on a spatial tessellation \cite{xie2020urban, zheng2014urban}.
It is a crucial problem given its impact on several aspects of society, from public safety \cite{zheng2014urban} to the definition of on-demand services \cite{zhou2018refined}, the management of land use \cite{jayarajah2018understanding}, and traffic optimization \cite{zheng2014urban}. For example, crowd flow predictors may help city managers and policymakers discover the traffic congestions in the city; people in business find potential areas of business investment; citizens improve travel plans and stagger the peaks of travel. 
These predictors may also help prevent or mitigate dangerous situations, such as creating massive crowds of people streamed into a strip region, by sending out warnings or evacuating people in advance. 

{\color{black} Crowd flow prediction is challenging because it requires dealing with both spatial and temporal dependencies.
Indeed, a region's out-flow may affect the in-flows of both near and far regions. 
At the same time, crowd flows are characterized by temporal closeness, trends, and periodicity. 
Temporal closeness marks the dependencies between events that are close in time; trends highlight patterns that repeat over time (e.g., weekends and working days); periodicity captures the repetitive nature of relevant events (e.g., rush hours in the morning).
Furthermore, exogenous factors such as weather conditions, holidays, and the presence of public city events may affect crowd flow patterns.  }

\indent {\bf Problem Definition. }
Given an individual's trajectory $T_u$ and a spatial tessellation $\mathcal{G}$ of the geographic space (generally a $i \times j$ grid), the set of locations (tiles) the trajectory intersects in a time interval $\Delta t$ is:
\begin{equation}
\small
q^t_{T_u} = \{ (p_k \rightarrow t) \in \Delta t \wedge (p_k \rightarrow (x,y)) \in (i,j) | (i,j)\},
\label{eq:intersect}
\end{equation}
where $(i,j)$ indicates a location on $\mathcal{G}$ and $p_k$ is the user $u$'s current location, identified by the coordinates $(x, y)$.
Let $Q$ be the set of locations covered by all the individual trajectories and let $t-1$, $t$, and $t+1$ be three consecutive time spans,
the incoming flow $\text{in}_t^{(i, j)}$ to a location $(i,j)$ is the number of the individuals that are in $(i, j)$ at time $t$ but were not in $(i, j)$ at time $t-1$. Similarly, the outgoing flow $\text{out}_t^{(i, j)}$ to location $(i, j)$ is the number of individuals that are in $(i,j)$ at time $t$ and move to another location at time $t+1$:
\begin{equation}
\small
 \text{in}_t^{(i,j)} = \sum_{T \in Q} |\{ t > 1 | (i,j) \notin q_T^{t-1} \wedge (i,j) \in q_T^{t} \}|;
 \mbox{\quad}
 \text{out}_t^{(i,j)} = \sum_{T \in Q} |\{ t > 1 | (i,j) \in q_T^{t} \wedge (i,j) \notin q_T^{t+1} \}|.
 \label{eq:pdef}
\end{equation}
We can represent the flows of a region as a tensor $X_t \in R^{2 \times I \times J}$, where one dimension is associated with the in-flow $(X_t)_{1,i,j} = \text{in}_t^{(i,j)}$ and the other with the out-flow $(X_t)_{2,i,j} = \text{out}_t^{(i,j)}$. 
Therefore, crowd flow prediction is the task of predicting $X_{t + \Delta}$ given the historical flows $\{ X_t | 1, \dots, X_t | t \}$. 
In most of the selected papers, $\Delta =1$. 
When $\Delta > 1$ (e.g., in \cite{wang2020seqst, zhou2019st}), the problem is named multi-step crowd flow prediction. 

A variant of crowd flow prediction aims at forecasting the entire origin-destination matrix (i.e., flows among pairs of locations) given the historical observations of crowd flows \cite{rong2019deep}.

\indent {\bf DL vs Traditional approaches.}
Crowd flow prediction may be tackled using classic time-series prediction models based on autoregression (AR), such as the AutoRegressive Moving Average (ARMA) \cite{box2015time}, the Autoregressive Integrated Moving Average (ARIMA) \cite{moorthy1988short, lee1999application}, and variants like the Stationary and Seasonal ARIMA (SARIMA) \cite{williams2003modeling}, vector ARMA \cite{kamarianakis2003forecasting} and space-time ARIMA \cite{kamarianakis2005space}. 
Since ARMA and ARIMA can make predictions only out of stationary time-series that do not statistically change over time, they cannot accurately predict new events. 
SARIMA, vector ARMA and space-time ARMA are designed to overtake this assumption, but they present other issues. 
For instance, SARIMA's predictive performance decreases when dealing with patterns with long seasonal periods. 
In general, the AR predictors have difficulties handling spatial dependencies and short-term samples, capturing patterns way before in the time series, and including additional features (e.g., weather conditions, presence of public city events),
making them ineffective for crowd flow prediction. 
\begin{figure}[!htb]
	\centering
    \includegraphics[width=1\textwidth]{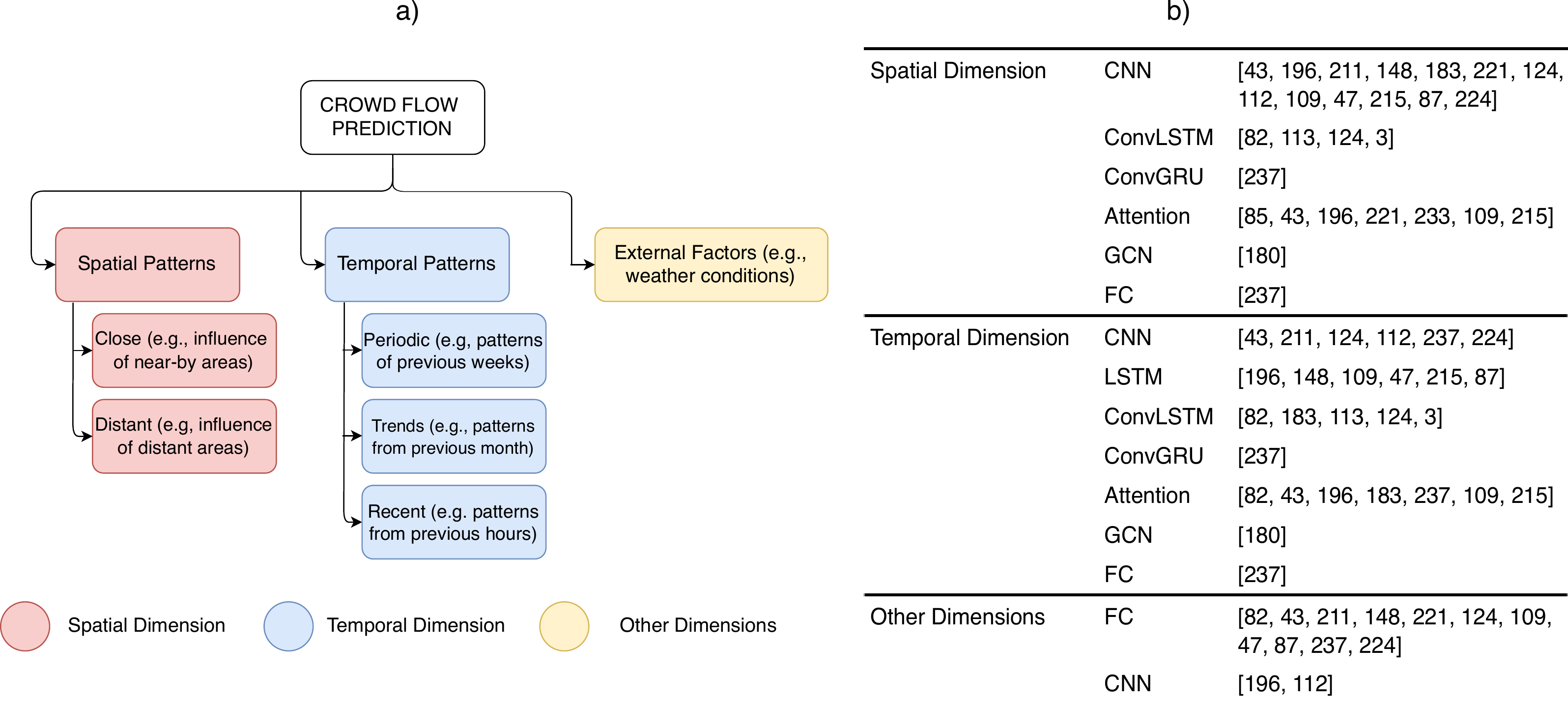}
\caption{\textbf{a)} The aspects crowd flow predictors should capture regarding the spatial dimension (red), the temporal dimension (blue), and the social and geographic dimensions (yellow) of human mobility data. 
\textbf{b)} DL modules that allow capturing each dimension, with reference to the selected papers in the literature that implement these modules.}
	\label{fig:crowd_flow_specialization}
\end{figure}

{\color{black}
In contrast, DL approaches can effectively model spatial and temporal dependencies and capture the influence of external factors (see Figure \ref{fig:crowd_flow_specialization}).
Since spatial tessellations describe a bidimensional space, DL predictors represent crowd flows as matrices and exploit the effectiveness of CNNs in detecting near and far away spatial and temporal dependencies in matrices.
At the same time, since gated RNNs such as LSTMs and GRUs may be used to model complex sequential patterns in the data, DL approaches may efficiently capture patterns in the temporal evolution of crowd flows.
CNNs and RNNs may be combined in ConvLSTM models, which may capture spatial and temporal patterns at the same time. 
Finally, we should also consider other aspects that may affect crowd flows, such as weather conditions and the presence of public events in the city. 
Such external factors may be handled with FCs and combined with the output of the CNN/RNN modules. 

}

{ \color{black} \indent{\bf Datasets and Evaluation metrics.} The most used datasets to train and evaluate crowd flow predictors are the Citi Bike dataset \cite{bikenyc}, which describes the trips between bike-sharing stations in New York City, and Zhang et al.'s dataset \cite{zhang2017deep}, which provides the aggregated incoming and outgoing flows for each tile of a squared tessellation of New York City and Beijing, extracted from raw taxis' GPS traces. 
Some papers use a dataset about the bike-sharing system of Washington D.C. \cite{bikewo}, 
and a dataset describing the pick-up and drop-off locations of taxis in New York City \cite{tlc}. 
We refer to Appendix \ref{app:gps} and \ref{app:social_media} for details about these datasets.

The performance of crowd flow predictors is evaluated as the error between the empirical  crowd flows and the predicted ones.
Commonly used error metrics are Mean Absolute Error (MAE), Root Mean Squared Error (RMSE), and Mean Absolute Percent Error (MAPE) (see Appendix \ref{app:error_metrics} for definitions and details).
}

\indent {\bf DL approaches. }

{ \color{black} 

Table \ref{tab:crowd} shows the selected papers on crowd flow prediction, highlighting the DL modules, evaluation metrics, and datasets they use, with a link to an implementation (if available).
}

\begin{table}[]
    \scriptsize
    \begin{tabularx}{\textwidth}{ c c c c c c c c X}
        \toprule
       & \scriptsize \textbf{Reference}   &   \scriptsize \textbf{Name} & \textbf{Year}       &   \scriptsize \textbf{DL Modules} &   \scriptsize \textbf{Evaluation}       &   \scriptsize \textbf{Dataset} &   \scriptsize \textbf{Code (https://bit.ly)} \\ \midrule 
        
        
        
        \multirow{19}{*}{\rotatebox[origin=c]{90}{\bf Crowd Flow Prediction}} & {\color{black} \emph{Jiang et al.}\cite{jiang2021deepcrowd} }  & DeepCrowd  & 2021  &  ConvLSTM, Attention, FC   & MSE, MAE, MAPE, RMSE & \cite{jiang2021deepcrowd} & \url{DeepCrowd} \\
        
        & {\color{black} \emph{Dai et al.}\cite{dai2021attention} }  & -  & 2021  &  Attention, CNN, FC &  RMSE & \cite{zhang2017deep} & - \\

        & {\color{black}\emph{Wang et al.}\cite{wang2020seqst} }  & SeqST   & 2020  &  LSTM, CNN, Attention  & MAE, RMSE & \cite{bikenyc, tlc} & - \\
        
        & {\color{black}\emph{Yang et al.}\cite{yang2020spatio} }  & ST-ESNet   & 2020  & CNN, FC  & RMSE & \cite{zhang2017deep} & - \\

        & \emph{Ren et al.}\cite{ren2020hybrid}   & HIDLST  & 2020  & LSTM, CNN, FC  & RMSE  & \cite{zhang2017deep}  & - \\
        & \emph{Tian et al.}\cite{tian2020deep}   & LDRSN  & 2020  & CNN, ConvLSTM, Attention   & RMSE, MAPE, MAE  & \cite{bikenyc, tlc}  & - \\
        & \emph{Yuan et al.}\cite{yuan2020deep}   & MV-RANet  & 2020  & CNN, Attention, FC  & RMSE, MAPE  &  \cite{zhang2017deep} & - \\
        & \emph{Liu et al.}\cite{liu20120dynamic}   & ATFM  & 2020  & ConvLSTM  & RMSE  & \cite{zhang2017deep} & \url{ATFM-2} \\

    & \emph{Sun et al.}\cite{sun2019predicting}   & MVGCN  &  2020 & GCN, FC  & RMSE, MAE  & \cite{tlc, zhang2017deep, bikewo, bikenyc} & - \\
         

         & {\color{black}\emph{Mourad et al.}\cite{mourad2019astir} }  & ASTIR  & 2019  & ConvLSTM, CNN, FC & RMSE & \cite{zhang2017deep} & \url{ASTIR_Model} \\
         
        & \emph{Zhou et al.}\cite{zhou2019st}   &  ST-Attn & 2019  &  Attention, FC & RMSE  & \cite{zhang2017deep, bikenyc}  & \url{ST-Attn} \\

        & {\color{black}\emph{Rong et al.}} \cite{rong2019deep} &  & 2019  & CNN, ResNet & RMSE, MAE  & - & - \\ 
         
        & \emph{Li et al.}\cite{li2019densely}   & ST-DCCNAL  & 2019  & CNN, Attention, LSTM, FC  & RMSE  & \cite{zhang2017deep}    & \url{ST-DCCNAL} \\
        
        & \emph{Lin et al.}\cite{lin2019deepstn}   & DeepSTN+  & 2019  & CNN  & RMSE, MAE  & \cite{bikenyc}  & \url{DeepSTN} \\
        & \emph{Du et al.}\cite{du2019deep}   & DST-ICRL  & 2019  & CNN, LSTM, FC  & RMSE, MAE  & \cite{zhang2017deep}  & \url{DST-ICRL} \\ 
        
        & \emph{Ai et al.}\cite{ai2019deep}   & -   & 2018  & ConvLSTM  & RMSE, MAE  & - & - \\
        & \emph{Yao et al.}\cite{yao2019revisiting}   & STDN  & 2018  & CNN, LSTM, Attention, FC  & RMSE, MAPE  & \cite{bikenyc}  & \url{STDN-2} \\
        & \emph{Jin et al. }\cite{jin2018st}   &  STRCN & 2018  & CNN, LSTM, FC  & RMSE  &  \cite{zhang2017deep}  & - \\
        & \emph{Zonoozi et al.}\cite{zonoozi2018periodic}   & PCRN   &  2018  & ConvGRU, CNN, FC & RMSE  & \cite{zhang2017deep}  & - \\

        
        & \emph{Zhang et al.}\cite{zhang2017deep}   & ST-ResNet  & 2017   & CNN, FC   & RMSE   & \cite{zhang2017deep}  & \url{ST-ResNet} \\

    \bottomrule    
     \end{tabularx} 
     \caption{List of selected papers tackling crowd flow prediction. 
     For each paper, we describe the name of the corresponding proposed model (if any), the year of publication, the DL modules used in the proposed solution, the metrics used for performance evaluation, the link to the public datasets used for training and testing the model, and the link to a repository with the code (if any). 
     The papers are sorted by year of publication in decreasing order.}
     \label{tab:crowd}
\end{table}

In their seminal work, Zhang et al. propose ST-ResNet \cite{zhang2017deep}, which consists of three modules that rely on CNNs to capture trends, periodic patterns, and temporal closeness. 
The modules' output is combined with the output of an FC that deals with external factors such as weather conditions and presence of public events. 
The model is evaluated on the datasets of taxis in Beijing and bikes in New York City \cite{zhang2017deep} using RMSE.

ST-ESNet \cite{yang2020spatio} extends STRes-Net \cite{zhang2017deep} by adding convolutional layers to upsample and downsample the matrices representing crowd flows.
The model is evaluated with the datasets of taxis in Beijing \cite{zhang2017deep} using RMSE. 

Many works in the literature combine CNNs with RNNs to exploit the latter's capability to deal with temporal patterns. 
STRCN (Spatio-Temporal Recurrent Convolutional Network) \cite{jin2018st} uses three CNNs to capture close, short-term (daily influence), and mid-term (the difference between workdays and weekends) spatial patterns.
The output of the CNNs is fed into three LSTMs, which handle temporal dynamics. 
STRCN also uses external features, such as weather conditions and features to distinguish between workdays and holidays, through an FC. 
The output of the FC and the LSTMs are combined. Model's performances are evaluated on the datasets of taxis in Beijing and bikes in New York City \cite{zhang2017deep} using RMSE.

STDN (Spatial-Temporal Dynamic Network) \cite{yao2019revisiting} consists of two CNNs: the first one captures the local spatial dependencies based on the similarities of historical traffic volumes to uncover the flows; the second CNN captures the traffic flows. 
The outputs of the two CNNs are pairwise multiplied and summed with an FC's output that handles external features (e.g., weather events). 
Moreover, STDN uses three LSTMs to capture the temporal dependencies of historical data describing the crowd flows. 
CNNs and LSTMs work on two separate pipelines, the outputs of which are summed together and forwarded to another LSTM with an attention mechanism that analyzes the temporal dynamics of the current day. Finally, the output of this latter LSTM is passed to an additional FC to perform crowd flow and traffic prediction.
The model performance is evaluated on the New York City's bikes dataset \cite{bikenyc} and on a dataset of taxis' flows in New York City \cite{tlc} using MAPE.

DST-ICRL (Deep Spatio-Temporal with Irregular Convolutional Residual Network)  \cite{du2019deep} combines convolutional residual units with LSTMs to capture the irregular properties of traffic flows in different transportation lines.
Similarly to STDN \cite{yao2019revisiting} and ST-RESNET \cite{zhang2017deep}, DST-ICRL uses three pipelines to uncover daily, weekly, and recent spatio-temporal dynamics and handles external features using an FC. 
The authors evaluate DST-ICRL on two private datasets of checkins from e-cards on buses and subways in Beijing, 
using MAE and RMSE. 

Another category of crowd flow predictors exploits attention mechanisms. 
ST-DCCNAL (Spatio-Temporal  Densely Connected Convolutional Networks and Attention LSTM) \cite{li2019densely} combines a CNN with an attentional LSTM to simplify the selection of the inputs.
The authors use DenseNet \cite{8099726} to cope with the spatial patterns and FCs to deal with the external features.
The outputs of DenseNet and the FCs are fed into an attentional LSTM to extract temporal patterns and make the prediction. The performances are evaluated on the datasets of taxis in Beijing and bikes in New York City \cite{zhang2017deep} using RMSE.

MV-RANet (Multi-View Residual Attention Network) \cite{yuan2020deep} uses two pipelines to deal with spatio-temporal patterns and mobility patterns. 
The first pipeline models the closeness, the periods, and the trends of crowd flows using three attention residual networks, the outputs of which are fused and passed to a convolutional layer.
The output of the convolutional layer is then passed to an FC. 
The second pipeline captures the mobility patterns in the data, generating three graphs: one for the transition probability, one for the transition distance, and one that mimics the flow patterns. 
The three graphs are then encoded using Node2Vec \cite{node2vec}, an algorithmic framework that learns a continuous feature representation of the nodes of a graph, and passed to an FC to extract relevant mobility patterns. 
The outputs of the FC handling the graph representation and the FC that handles the spatio-temporal patterns are fed into an additional FC responsible for making the prediction. 
MV-RANet is evaluated using RMSE, MAPE, and MAE on the flows of taxis in Beijing \cite{zhang2017deep} 
and a private mobile phone data (CDRs) for the city of Sanya.

LDRSN (Local-Dilated Region-Shifting Network) \cite{tian2020deep} consists of five modules. 
The first one is a CNN that handles local spatial dependencies; the second one uses dilated units to deal with distant spatial dependencies. 
A $k$-dilated unit is a convolution that captures the correlation of two regions with a distance of $k$. 
The outputs of these two modules are fused and forwarded to two other modules: one handles long-term temporal dependencies, the other deals with short-term temporal patterns. 
The long-term module consists of three pipelines with consecutive convolutional layers, each followed by attention LSTMs. 
A single ConvLSTM handles the short-term patterns. 
The outputs of the two temporal modules are fed into a final module: as a first step, it sums the outputs of the two temporal modules and forwards the result to a convolutional layer to make the prediction.
The model is evaluated on the dataset of bikes' \cite{zhang2017deep} 
and taxis' in New York City \cite{tlc} 
using MAE, MAPE, and RMSE.

To capture long-range spatial dependencies, DeepSTN+ (Deep Spatio Temporal Network Plus) \cite{lin2019deepstn} replaces the traditional residual units and the convolutional layers with ResPlus and ConvPlus. 
ResPlus units employ a ConvPlus layer and a standard convolutional layer to capture distant spatial dependencies. 
The idea beyond a ConvPlus layer is to separate the channels of the input matrix and use an FC to capture the long-range spatial dependencies among each pair of regions. 
Finally, DeepSTN+ uses an average pooling layer before the FC to reduce the number of parameters.
Another peculiarity of DeepSTN+ is that it uses temporal factors and the distribution of POIs to gain prior knowledge of the crowd flows. 
The experiments are conducted on the dataset of bikes in New York City \cite{bikenyc} 
and a private dataset from the most popular social network vendor in China 
using MAE and RMSE.

{\color{black} ST-Attn \cite{zhou2019st} is specifically designed for multi-step crowd flow prediction. 
It relies on an encoder and a decoder based on two attention mechanisms (one capturing spatial patterns and one capturing temporal patterns) and several FCs.
The encoder is a stack of a spatial and temporal attention mechanism and FC layers. 
The decoder has two consecutive attention mechanisms to capture spatial and temporal patterns: the first one uses historical observations about crowd flows, the second one uses the output of the first attention mechanism and the output of the encoder.
The decoder's output is passed to an FC layer to predict the $\Delta$ next crowd flows. 
The model is evaluated for $\Delta \in 1 \dots 6$ on the datasets of taxis in Beijing \cite{zhang2017deep}, taxis in New York City \cite{tlc}, and bikes in New York City \cite{bikenyc} using RMSE.  

ASTIR \cite{mourad2019astir} consists of three pipelines each using ConvLSTMs to capture periodic, long, and short-terms spatial and temporal dependencies, respectively. 
The outputs of the three ConvLSTMs are passed to three attention mechanisms, whose outputs are passed to three ConvLSTMs. 
The outputs of the three pipelines are combined with the results of an FC handling external factors and passed to a tanh activation function to perform the prediction. 
ASTIR is evaluated on the dataset of taxis in Beijing and bikes in New York City \cite{zhang2017deep} using RMSE.

SeqST \cite{wang2020seqst} performs multi-step crowd flow prediction using a sequence to sequence GAN. 
The GAN's generator and discriminator rely on CNNs and LSTMs to extract spatio-temporal patterns from the data. 
In the generator, they use further CNNs to consider the impact of external factors, while the discriminator drives the generator to predict realistic crowd flows. 
The model is evaluated on the dataset of taxis \cite{tlc} and bikes in New York City \cite{bikenyc} using MAE and RMSE.

DeepCrowd \cite{jiang2021deepcrowd}uses three pipelines to capture patterns in previous hours, days, and weeks. 
It consists of three ConvLSTMs that downsample and upsample the input data. 
Each convLSTM is followed by an attention mechanism to extract relevant local spatio-temporal patterns.
The outputs of the three attention mechanisms are concatenated and passed to an additional attention mechanism to further extract relevant patterns. 
The output of the attention mechanism is passed to a CNN to predict crowd flows. 
DeepCrowd is evaluated on a public dataset of crowd flows in Tokyo and Osaka, Japan \cite{jiang2021deepcrowd} using MSE, MAE, MAPE, and RMSE.
}

In contrast with the majority of solutions to crowd flow prediction, Dai et al. \cite{dai2021attention} uses only a single pipeline to capture both near and far away temporal and spatial patterns. All the matrices representing crowd flows are passed to a temporal attention mechanism, whose output is passed to a spatial attention mechanism. 
The output of the spatial attention mechanism is fed into a CNN to extract spatio-temporal patterns.
The output of this CNN is fused with the results of a FC which handles  external features. 
The model is evaluated on the dataset of taxis in Beijing and bikes in New York City \cite{zhang2017deep} using RMSE.

Some other models address slightly different definition of crowd flow prediction \cite{zonoozi2018periodic,ai2019deep,sun2019predicting}.
PCRN (Periodic Convolutional Recurrent Networks) \cite{zonoozi2018periodic} solves the problem of predicting the distribution of presences in a city.   
The authors use a pyramidal model made of three ConvGRUs (which has both the advantages of CNNs and GRUs) with an external module that captures the patterns' periodicity by memorizing the periodic representations learned by the stacked ConvGRUs. 
The authors propose three ways to retrieve and update the periodic patterns: using a sequential periodic representation, using an estimated average of periodic representation, adopting a temporally ordered representation. The outputs of the periodic module and the ConvGRUs are fused. The model is evaluated on the datasets of taxis in Beijing and bikes in New York City \cite{zhang2017deep} using RMSE.

Ai et al. \cite{ai2019deep} aim to predict the short-term distribution of features that describe the movements of bikes of a dockless bike-sharing system. 
They rely on ConvLSTMs to address the spatial and temporal dependencies. 
ConvLSTMs take as input a spatio-temporal sequence composed of the number of bicycles in an area, the distribution uniformity, the usage distribution, and the time of the day to predict their values in the near future.
The authors evaluate their model using private datasets from two bike-sharing companies in Chengdu, China. 
The metrics used for the evaluation are MAE and RMSE. 

Some recent works solves the problem of flow prediction, a variant of crowd flow prediction in which they aim to use historical observations of crowd flows to forecast the entire origin-destination matrix, i.e., flows among pairs of locations. 

MVGCN (Multi-View Graph Convolutional Network) 
\cite{sun2019predicting} does so handling the external features (e.g., weather data) with two FCs, which deals with weather information and meta-information such as time and day of the week. 
The authors select key time snapshots to process graphs representing recent, daily, weekly, monthly and quarterly mobility flows. Each node in these graphs represents a region with time-varying flows. The graphs are forwarded to five graph convolutional networks and the outputs of the seven networks (five graph convolutional networks and two FCs for the external features) are fused using a multi-view fusion mechanism. 
The fusion module's output is fed into an additional FC that outputs the predicted graph, in which each node is a region and the links are the predicted flows. 
The experiments are conducted on the dataset of bikes \cite{bikenyc} and taxis \cite{tlc} in New York City, bikes in Washington D.C. \cite{bikewo}, 
and taxis in Beijing \cite{zhang2017deep}. 
In both cases, RMSE and MAE are used to evaluate the model's performance. 

{\color{black}
Rong et al. \cite{rong2019deep} use a distinct CNN for each location represented as an image-like matrix, to extract long and short-terms spatial and temporal features. 
Each matrix is made of several channels, each corresponding to one time interval of the historical observations.
The model is evaluated on a private dataset of a Chinese social network referring to flows in Beijing using MAE and RMSE.
}
\section{Generative Models}
\label{sec:generative_models}

Generative models of human mobility aim at generating realistic spatio-temporal trajectories (trajectory generation, Section \ref{sec:task_generative}) or mobility flows (flow generation, Section \ref{sec:flow_generative}).

In this Section, we describe both tasks, discussing how DL brings significant improvements with respect to traditional approaches, and describing the relevant state-of-the-art solutions to each task, with a reference to the public datasets and the metrics used for training and testing the models.

\subsection{Trajectory Generation}
\label{sec:task_generative}

The goal of generative models of individual human mobility is to generate synthetic trajectories with realistic mobility patterns \cite{karamshuk2011human, hess2015data, wang2019urban, barbosa2018human, feng2020learning, shin2020user}.
The generated synthetic trajectories must reproduce a set of spatial and temporal mobility patterns, such as the distribution of characteristic distances traveled and the predictability of human whereabouts (see Appendix \ref{app:mobility_patterns}). 
The use of generative mobility models is crucial in many applications.
First, synthetic trajectories are useful to the performance analysis of networking protocols such as mobile ad hoc networks, where the displacements of network users are exploited to route and deliver the messages \cite{karamshuk2011human, hess2015data, tomasini2017effect}. 
Second, synthetic trajectories are fundamental for urban planning, what-if analysis, and computational epidemiology, e.g., simulating changes in urban mobility in the presence of new infrastructures, epidemic diffusion, terrorist attacks, or international events \cite{cutter2013disaster, unisdr2012disaster, yabe2016framework, yabe2017cityflowfragility}. 
Furthermore, generative models are a viable solution to protect geo-privacy of trajectory data \cite{mir2013dpwhere, fiore2019privacy, pellungrini2020modeling}: while disclosing real data requires a hard-to-control trade-off between uncertainty and utility, synthetic records that preserve statistical properties may achieve in multiple tasks comparable performance to real data. 

{\color{black} Solving trajectory generation requires capturing, simultaneously, the temporal and spatial patterns of individual human mobility.
A realistic generative model should reproduce the temporal statistics observed empirically, including the number and sequence of visited locations together with the time and duration of the visits. 
In particular, the biggest hurdle consists of the simultaneous description of an individual's routine and sporadic out-of-routine mobility patterns. 
Regarding spatial patterns, a generative model should reproduce the tendency of individuals to move preferably within short distances \cite{gonzalez2008understanding, pappalardo2013understanding}, the heterogeneity of characteristic distances \cite{gonzalez2008understanding, pappalardo2013understanding} and their scales \cite{alessandretti2020scales}, the tendency of individuals to split into returners and explorers \cite{pappalardo2015returners}, the routinary and predictable nature of human displacements \cite{song2010limits}, and the fact that individuals visit a number of locations that are constant in time \cite{alessandretti2018evidence}.
}

\indent {\bf Problem Definition. } 
A generative mobility model $M$ is any algorithm able to generate a set of $n$ synthetic trajectories $\mathcal{T}_M = \{ T_{a_1}, \dots, T_{a_n} \}$, which describe the movements, during a certain period of time, of $n$ independent agents $a_1, \dots, a_n$. 
The synthetic trajectory generated for a single agent $a_i$ should be in the form of Definition \ref{def:trj_simple}, i.e., a time-ordered sequence  $T_{a_i}=\langle p_{1}, p_{2}, ..., p_{k} \rangle$ composed by spatio-temporal points, describing the $k$ locations visited by $a_i$.
The realism of $M$ is evaluated with respect to:
\begin{enumerate}
\item A set of spatial patterns ($s_1, \dots, s_{m_s}$) and temporal patterns ($t_1, \dots, t_{m_t}$) $\mathcal{K} = \{ s_1, \dots, s_{m_s}, t_1, \dots, t_{m_t}\}$ (see Appendix \ref{app:mobility_patterns}). 
The patterns refer to the distributions of individual measures, which quantify aspects related to the mobility of a single individual (e.g., radius of gyration, mobility entropy), or collective measures, which quantify aspects related to the mobility of a region as a whole (e.g., OD matrices). 
A realistic $\mathcal{T}_M$ is expected to reproduce as many mobility patterns as possible.

\item A set $\mathcal{X} = \{T_{u_1}, \dots, T_{u_m} \}$ of real mobility trajectories corresponding to $m$ real individuals $u_1 \dots u_m$ that move on the same region as the one on which synthetic trajectories are generated. 
Generally, a portion $\mathcal{X}_{\mbox{\footnotesize train}} \in \mathcal{X}$ is used to train $M$ or to fit its parameters. The remaining part $\mathcal{X}_{\mbox{\footnotesize test}}$ is used to compute the set $\mathcal{K}$ of patterns, which are compared with the patterns computed on $\mathcal{T}_M$.

\item A function $D$ that computes the dissimilarity between two distributions, such as the KL divergence or the JS (see Appendix \ref{app:divergence} for definitions). 
Specifically, for each measure in $f \in \mathcal{K}$, $D(P_{(f, \mathcal{T}_M)}||P_{(f, \mathcal{X}_{\mbox{\footnotesize test}})})$ indicates the dissimilarity between $P_{(f, \mathcal{T}_M)}$, the distribution of the measures computed on the synthetic trajectories in $\mathcal{T}_M$, and $P_{(f, \mathcal{X}_{\mbox{\footnotesize test}})}$, the distribution of the measures computed on the real trajectories in $\mathcal{X}_{\mbox{\footnotesize test}}$.
The lower $D(P_{(f, \mathcal{T}_M)}||P_{(f, \mathcal{X}_{\mbox{\footnotesize test}})})$, the more realistic model $M$ is with respect to $f$ and $\mathcal{X}_{\mbox{\footnotesize test}}$.
\end{enumerate}

\indent {\bf DL vs Traditional approaches.} 
There is a vast literature on mechanistic generative models that reproduce simple temporal, spatial, and social patterns of human mobility \cite{barbosa2018human, pappalardo2017data, karamshuk2011human, hess2015data, wang2019urban}.
For example, in the Exploration and Preferential Return (EPR) model \cite{song2010modelling}, an agent can choose between two competing mechanisms: exploration, during which an agent chooses a new location never visited before, based on a random walk process with a power-law jump-size distribution; and preferential return, in which an agent returns to a previously visited location based on its frequency.
Several studies subsequently improved the EPR model by adding increasingly sophisticated spatial or social mechanisms \cite{pappalardo2015returners, barbosa2015effect, alessandretti2018evidence, toole2015coupling, cornacchia2021stsepr}.
EPR and its extensions focus mainly on the spatial aspects of human mobility, implementing unrealistic temporal mechanisms.
TimeGeo \cite{jiang2016timegeo} and DITRAS \cite{pappalardo2017data} improve the temporal mechanism integrating into an EPR-like model a data-driven model that captures the circadian propensity to travel and out-of-routine trips.
Although mechanistic models have the advantage of being interpretable by design, their realism is limited because of the simplicity of the implemented mechanisms.

\begin{figure}[!htb]
	\centering
    \includegraphics[width=1\textwidth]{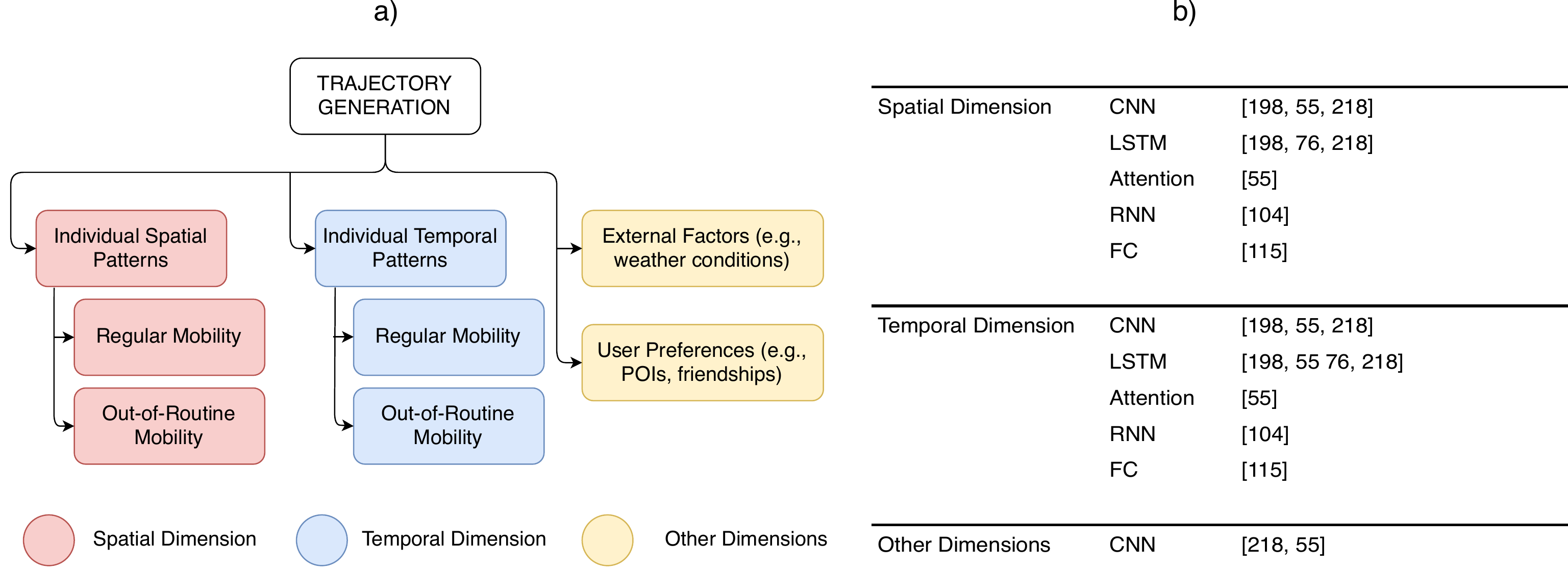}
\caption{\textbf{a)} The aspects trajectory generators should capture regarding the spatial dimension (red), the temporal dimension (blue), and the social and geographic dimensions (yellow) of human mobility data. 
\textbf{b)} DL modules that allow to capture each dimension, with the reference to the selected papers in the literature that implement these modules.}
	\label{fig:generative_specialization}
\end{figure}
The limitations mentioned above can be tackled using DL generative paradigms such as GANs and VAEs. In both cases, models rely on DL modules to learn the distribution of data and generate mobility trajectories coming from the same distributions (Figure \ref{fig:generative_specialization}). 
Thanks to the versatility of DL modules, GANs and VAEs can capture different aspects simultaneously (e.g., spatial, temporal, social dimensions of mobility) while traditional approaches can capture peculiar aspects of mobility only (e.g., only the spatial dimension). Moreover, DL models can capture complex and non-linear relationships in the data that traditional approaches may fail to capture. Therefore, such models can generate more realistic data than traditional models.

\indent \textbf{Datasets and evaluation metrics.} 
{\color{black}
Trajectory generators are commonly trained and evaluated on GPS datasets such as Geolife \cite{zheng2010geolife}, MDC  \cite{laurila2012mobile}, and taxi traces in San Francisco \cite{epfl-mobility-20090224}.
Additional information about these datasets can be found in Appendix \ref{app:gps}. 
The distance between the distribution measures the performances of trajectory generators (e.g., using KL or JS divergence, see Appendix \ref{app:divergence} for details) of standard mobility metrics computed on a real dataset and the generated dataset. 
Appendix \ref{app:distance_metrics} describes a set of standard mobility metrics commonly used to evaluate the realism of trajectory generators.}

\indent {\bf  DL approaches.} 
\begin{table}[]
    \scriptsize
    \begin{tabularx}{\textwidth}{ c c c c c c c c X}
        \toprule
       & \scriptsize \textbf{Reference}   &   \scriptsize \textbf{Name} & \textbf{Year}       &   \scriptsize \textbf{DL Modules} &   \scriptsize \textbf{Evaluation}       &   \scriptsize \textbf{Dataset} &   \scriptsize \textbf{Code (https://bit.ly)} \\ \midrule 

        
        
        \multirow{12}{*}{\rotatebox[origin=c]{90}{\bf Trajectory Generation}} & { \color{black}Wang et al. \cite{wang2021large} } & TSG & 2021 & GAN, CNN, LSTM & \makecell{ distances, $r_g$, $p(r, d)$} & \cite{moreira2013predicting} & \url{TSG_Model} \\

        & Feng et al. \cite{feng2020learning} & MoveSim & 2020  & \makecell{GAN, self-attention \\ CNN} & \makecell{ distances, $r_g$, $p(r, d)$, \\ DailyLoc, G-rank, I-rank} &  \cite{zheng2010geolife} & \\

        & Huang et al. \cite{huang2019variational} & SVAE & 2019  & VAE, LSTM & MDE & 
- & - \\

     & Ouyang et al. \cite{ouyang2018non} & Ouyang GAN & 2018  & WGAN, CNN & \makecell{ $p(r)$, $p(r, t)$, $p(r, d)$, \\ $p(r, d_{total})$,  $p_{d_{total}}$, \\ location frequency } & 
     \cite{laurila2012mobile} & - \\

    & Kulkarni et al. \cite{kulkarni2018generative} & - & 2018  & RNN, GAN & \makecell{visitation frequency, \\ statistical similarity, \\ privacy tests} & 
\cite{laurila2012mobile} & -\\

    & Yin et al. \cite{yin2018gans} & - & 2018 & GAN, FC & \makecell{reconstruction error, utility loss} & 
\cite{epfl-mobility-20090224} & - \\

    & Liu et al. \cite{liu2018trajgans} & trajGANs & 2018  & GANs & - & - & - \\
        
    \bottomrule    
     \end{tabularx} 
     \caption{List of selected papers tackling trajectory generation. 
     For each paper, we describe the name of the corresponding proposed model (if any), the year of publication, the DL modules used in the proposed solution, the metrics used for performance evaluation, the link to the public datasets used for training and testing the model, and the link to a repository with the code (if any). 
     Papers are sorted by year of publication in decreasing order.}
     \label{tab:generative}
\end{table}
Most of the DL approaches to trajectory generation are based on GANs, while a few are based on VAEs.
Table \ref{tab:generative} reports a selection of the most relevant DL approaches to trajectory generation.

In their vision paper, Liu et al. \cite{liu2018trajgans} propose the trajGANs framework to address the potential and challenges of using GANs for trajectory generation.
Similar to a typical GAN \cite{goodfellow2014generative}, a trajGAN consists of a generator $G$, which accepts a random vector $z$ and generates a dense representation of synthetic trajectories samples, and a discriminator $D$, which classifies an input trajectory sample into ``real'' or ``fake''.
Liu et al. \cite{liu2018trajgans} suggest the use of Recurrent Neural Networks (RNNs) to create dense representations of trajectories and transform between trajectories and distributional representations. 

Ouyang et al. \cite{ouyang2018non} represent a trajectory as a sequence of stays, each with a geographic location, start time and duration. 
Specifically, a trajectory is an $n_1 \times n_2 \times k$ matrix, where $n_1 \times n_2$ is the size of a squared tessellation, and $k$ is the maximum number of stay repetitions for each location (set in the experiments to $k=4$).
A Wasserstein GAN \cite{gulrajani2017improved} is used to train a trajectory generator. 
Both the generator and the discriminator are based on a CNN.
The experiments are conducted on the MDC dataset \cite{laurila2012mobile} using a $64 \times 64$ squared tessellation on the city of Lausanne, Switzerland. 
The similarity between synthethic and real trajectories is evaluated using the JS divergence on the popularity $p(r)$ and temporal popularity $p(r, t$) of locations, the staying patterns $p(r, d)$, the semantic importance $p_{d_{total}}(r)$, the semantic distance, and G-rank. 

Song et al. \cite{song2019generating} use a CNN with four layers within a GAN framework to generate trajectories represented as $512\times512$ matrices. 
In a data convolution process, the input $512\times512$ matrices are resized into a $32\times32$ matrix. 
Given the small size of the available (private) mobile phone datasets, the experiments include synthetic ones obtained by randomly shuffling the real trajectories.
In a deconvolution process, the GAN's output (a $32\times32$ matrix) is resized back to a $512 \times 512$ matrix using the nearest neighbor function. 
No quantitative evaluation of the model's realism is provided.

Huang et al. \cite{huang2019variational} present SVAE (Sequential
Variational Autoencoder), a generative model based on a combination of a Variational Autoencoder (VAE) and an LSTM, combining the ability of VAEs to construct a latent space that captures salient features of the training data with the ability of LSTMs to process sequential data.
In a data processing phase, they force the input 
trajectories to have fixed timestamps, and in the experiments they evaluate the realism of SVAE through the Mean Distance Error (MDE) between each step of real and synthetic trajectory pairs, finding that the reconstruction error of SVAE is smaller than 800 meters.

Kulkarni et al. \cite{kulkarni2018generative}  benchmark the performance of RNNs, SeqGAN \cite{yu2017seqgan}, RGAN \cite{esteban2017real} and nonparametric copulas to generate synthetic trajectories. 
They compare the generated trajectories with real ones extracted from the MDC dataset \cite{laurila2012mobile} based on geographic and semantic similarity, statistical similarity, long-range
dependencies and privacy tests.
They find that copulas have an advantage over all other methods in terms of both model performance and computational time.

MoveSim \cite{feng2020learning} is a model-free GAN framework that integrates the domain knowledge of human mobility regularity. 
The generator consists of a self-attention-based sequential model to capture the temporal transitions in human mobility. 
The discriminator consists of a mobility regularity-aware loss to distinguish the generated trajectory from a fake one. 
The mobility regularities of spatial continuity and temporal periodicity are used to pre-train the generator and discriminator to accelerate the learning procedure. 
They conduct experiments on a private mobile phone dataset, using the base station as the spatial unit, and on GeoLife \cite{zheng2010geolife}, projecting GPS coordinates into a grid. 
As for the temporal granularity set the basic time slot of a trajectory as half an hour of the day.
The realism of MoveSim is evaluated based on the distribution of distances, radius of gyration, number of locations visited daily, G-rank, and I-rank using the JS divergence with respect to real trajectories.

Yin et al. \cite{yin2018gans} use a GAN-based framework to generate density distributions rather than trajectories, i.e., the number of users in each location at each time slot. 
Both the generator and the discriminator are implemented with FCs.
Experiments are conducted on a dataset extracted from MoMo (spatial resolution 2km, time slots of 30 minutes) and on the dataset of taxis in San Francisco (squared tessellation of 50km$\times$50km, time slots of two minutes) \cite{epfl-mobility-20090224}.
The proposed model is evaluated in terms of reconstruction error and utility loss outperforms the differential privacy approach in data utility and attack error.

{
\color{black}

TSG (Two-Stage GAN) \cite{wang2021large} consists of two GANs with different objectives. The first GAN captures spatio-temporal patterns from the trajectories, mapped into a spatial tessellation leveraging CNNs (both for generator and discriminator). 
The generated matrices are then processed and the origin (enter point) and destination (exit point) of the trips are extracted. 
The second GAN extracts road information from road maps using CNNs, and uses LSTMs to generate a trajectory between an origin road and a destination road on the road network. 
The model performance is evaluated using the dataset of taxis in Porto \cite{moreira2013predicting} by measuring the JS divergence between the distribution of trajectory lengths and the frequencies of the top 50 visited places. 
}

{ \color{black} 
\subsection{Flow Generation}
\label{sec:flow_generative}
Flow generation consists of generating the flows between a set of geographic locations, given some locations' characteristics (e.g., population, POIs, land use, distance to other locations) and without any information about the real flows, \cite{barbosa2018human}.
Flow generation is crucial to many aspects of our society, such as transport planning \cite{erlander1990gravity} and spatial economics \cite{karemera2000gravity, patuelli2007network, prieto2018scaling} to reduce inequalities and to design more sustainable communities, and the modeling of epidemic spreading patterns \cite{balcan2010modeling, li2011validation, cevik2020going, zhang2020exploring}. 
Solving flow generation requires capturing the spatial patterns of close and distant flows, dependencies in the mobility network, and the characteristics of the locations.  

\indent \textbf{Problem definition.}
Given a tessellation $\mathcal{G}$ over a region $A$, the flow, $y(g_i, g_j)$, between locations $g_i$ and $g_j$ represents the number of people moving from $g_i$ to $g_j$. 
The total outflow, $O_i$, from location $g_i$ is the total number of people originating from location $g_i$, i.e., $O_i = \sum_j y(g_i, g_j)$. 
Flow generation aims to estimate $y(g_i, g_j) \forall$ $i, j \in \mathcal{G}$, $i \neq j$, i.e., the flow between each pair of locations $(g_i, g_j)$in given their total outflows $O_i$ and $O_j$. 

\indent \textbf{DL vs Traditional approaches.}
Flow generation has attracted interest for a long time ago. Notably, in 1946 George K. Zipf proposed a model to estimate mobility flows, drawing an analogy with Newton's law of universal gravitation \cite{zipf1946p}. 
This model, known as the gravity model, is based on the assumption that the number of travelers between two locations (flow) increases with the locations' populations while decreases with the distance between them \cite{barbosa2018human}.   
Despite the significant results achieved by the gravity model, it suffers from several drawbacks, including the inability to capture the structure of the real flows accurately and the more significant variability of real flows than expected \cite{simini2020deep}. Moreover, generations are done without considering other factors (e.g., POIs, street networks and others).
Other mechanistic approaches aim to extend the gravity model by capturing additional information like the radiation model.
All the models mentioned cannot capture non-linear relationships in the data and rely on a limited set of parameters. 

\begin{figure}[!htb]
	\centering
    \includegraphics[width=1\textwidth]{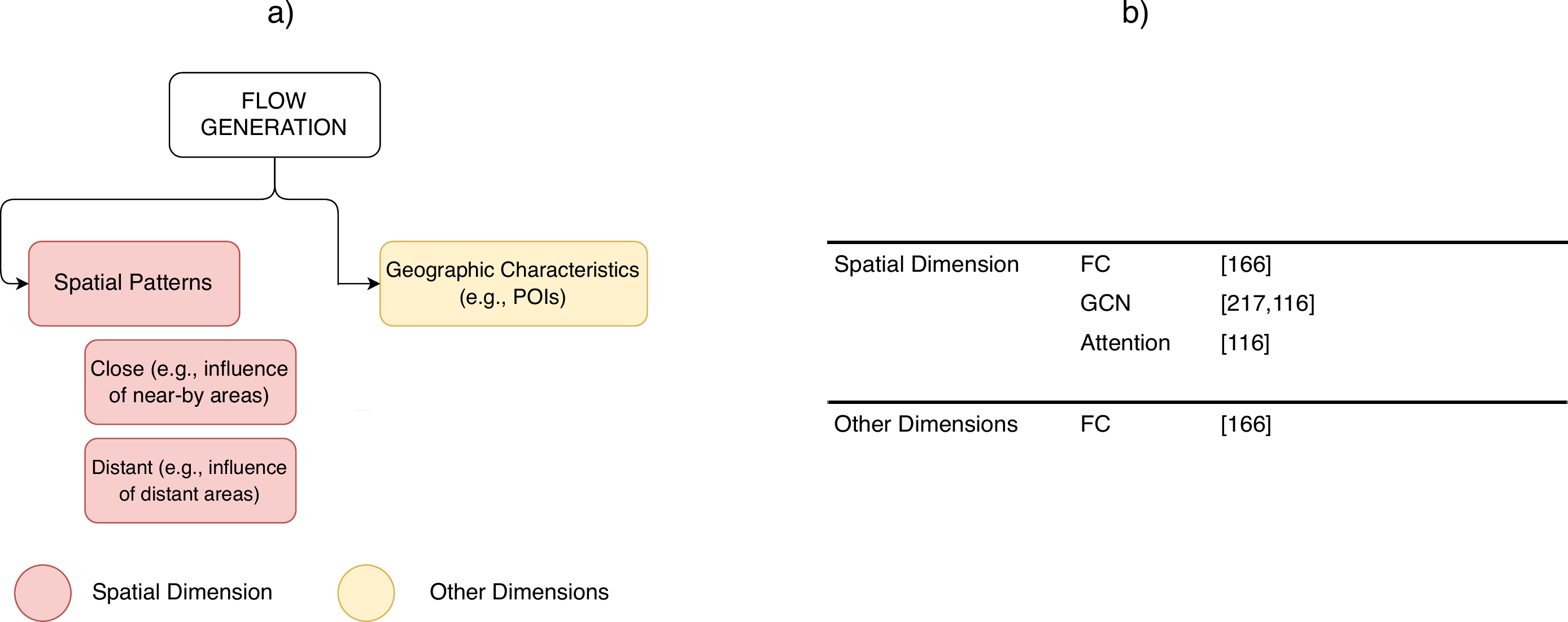}
\caption{\textbf{a)} The aspects flow generators should capture regarding the spatial dimension (red) and the social and geographic dimensions (yellow) of human mobility data. 
\textbf{b)} DL modules that allow to capture each dimension, with the reference to the selected papers in the literature that implement these modules.}
	\label{fig:flowgen_specialization}
\end{figure}

In contrast, DL approaches can capture complex and non-linear relationships in the data and easily integrate additional information about locations such as population, POIs, population (Figure \ref{fig:flowgen_specialization}). 
For example, an FC can be fed with data representing POIs and, similarly, POIs can be a node feature in a graph neural network. 
Finally, CNNs can capture spatial relationships that traditional approaches cannot capture. Similarly, FC networks can be fed with additional features representing areas near the origin and the destination of a flow to characterize its spatial patterns.

\indent \textbf{Datasets and evaluation metrics.}
Flow generators are generally evaluated on commuting data from censuses from official statistics institutes \cite{simini2020deep, kang2020multiscale}. 
An alternative dataset is the set of traces by taxis in Beijing \cite{zheng2011t-drive, yao2020spatial} aggregated into flows (see Appendix \ref{app:data} for details).
Flow generation is commonly evaluated as the CPC (Common Part Of Commuters) between real and generated flows.
Other metrics commonly used for this purpose are MAE, RMSE and MAPE. 
Additional information about these metrics can be found in Appendix \ref{app:error_metrics}.

\indent \textbf{DL approaches.}
There is a limited literature tackling flow generation (Table \ref{tab:flowgen}).

DeepGravity \cite{simini2020deep} use FCs to extend the original gravity model with the ability to capture non-linear relationships and the possibility to integrate additional information to characterize the locations easily (e.g., population, POIs). 
The model is evaluated on commuting data from official statistics in the UK and Italy and a dataset of flows aggregated from individual trajectories in the US \cite{kang2020multiscale} using CPC and RMSE.
The authors also provide meaningful explanations of the generated flows in terms of features of the origin and destination locations.

SI-GCN (Spatial Interaction GCN) \cite{yao2020spatial} consists of three parts: \emph{(i)} a layer for managing the spatial representation of the data (e.g., construct the local graph, negative sampling, features organization); followed by \emph{(ii)} an encoder that uses graph convolutions to generate a representation in a latent space for all the geographical units; and \emph{(iii)} a decoder in charge of generating the missing flows starting from the latent representation. SI-GCN is evaluated on the dataset of T-Drive \cite{zheng2011t-drive} using RMSE, MAPE and CPC as evaluation metrics.

GML (Geocontextual Multitask Embedding Learner) \cite{liu2020learning} captures the spatial correlation from geographic contextual information and relies on two graph neural networks with attention (GAT). The role of the GATs is to learn an embedding representation that is then passed to a gradient boosting that is in charge of generating the flows. 
The model is evaluated on commuting data regarding New York City using MAE, RMSE, and CPC.

\begin{table}[]
    \scriptsize
    \begin{tabularx}{\textwidth}{ c c c c c c c c X}
        \toprule
       & \scriptsize \textbf{Reference}   &   \scriptsize \textbf{Name} & \textbf{Year}       &   \scriptsize \textbf{DL Modules} &   \scriptsize \textbf{Evaluation}       &   \scriptsize \textbf{Dataset} &   \scriptsize \textbf{Code (https://bit.ly)} \\ \midrule 

        
        
        \multirow{3}{*}{\rotatebox[origin=c]{90}{\bf F. G.}} & Yao et al. \cite{yao2020spatial} & SI-GCN & 2020 & GCN & RMSE, MAPE, CPC & \cite{zheng2011t-drive} & \\
        
        & Simini et al. \cite{simini2020deep} & Deep Gravity & 2020 & FC & CPC & \cite{kang2020multiscale} & \href{https://bit.ly/DeepGravity}{DeepGravity} \\
        
        & Liu et al. \cite{liu2020learning} & GMEL & 2020 & GNN, Attention & RMSE, MAE, CPC &  & \href{http://bit.ly/GMEL-Code }{GMEL-Code} \\

    \bottomrule    
     \end{tabularx} 
     \caption{List of selected papers tackling flow generation. 
     For each paper, we describe the name of the corresponding proposed model (if any), the year of publication, the DL modules used in the proposed solution, the metrics used for performance evaluation, the link to the public datasets used for training and evaluating the model, and the link to a repository with the code (if any).
     The papers are sorted by year of publication in decreasing order.}
     \label{tab:flowgen}
\end{table}

}

\section{Conclusions}
\label{sec:conclusion}

In this survey, we proposed a perspective on DL approaches to human mobility, focusing on next-location prediction, crowd flow prediction, trajectory generation, and flow generation.
For each task, we highlighted the challenges related to solving it and how DL may help capture human mobility better than traditional models. We described a selection of relevant state-of-the-art solutions to each task.
As an additional contribution to the community, we created a GitHub repository (\href{https://bit.ly/DL4HM}{bit.ly/DL4HM}) where researchers can contribute to maintaining the list of relevant papers always up-to-date.
Our survey reveals that predictive tasks (next-location and crowd flow prediction) are well established in the community and addressed by a significant variety of DL approaches. 
In contrast, the usage of DL for generative tasks (trajectory generation and flow generation) is more recent and should require more attention in the future.

Our overview of the state of the art of DL for human mobility reveals that existing solutions suffer from several limitations, and many relevant aspects need to be addressed in the future.  
In particular, we identify the following open challenges.

\paragraph{Geographic Transferability.} 
Although DL models can capture complex mobility patterns automatically, they strictly depend on the data used for training and may not be \emph{geographically transferable}, i.e., one model trained on a specific region can be used to predict locations or crowd flows or to generate synthetic trajectories on a distinct, non-overlapping region.
Geographic transferability can be crucial in situations where there is a scarcity or even absence of mobility data for a region, and it poses several challenges related, for example, to the design of a suitable encoder of the mobility trajectories or flows. 
As a first tentative in this direction, RegionTrans \cite{wang2018crowd} provides insights on how transferability can be tackled for crowd flow prediction. Still, more work is needed to address this open challenge.

\paragraph{Explainability.}
DL models are by nature \emph{opaque}, i.e., they are black-boxes from which it is hard to reconstruct the reasoning that led to the generation of a trajectory or the prediction of a location or flow. 
Nonetheless, explainability is crucial for gaining a deeper understanding of mobility patterns and highlighting the presence of biases in the model's reasoning.
It is important to develop mobility-related explanations that provide examples and counter-examples to validate trajectories and crowd flows from different perspectives.
While models rely on many features, either external ones (e.g., weather data, POIs) or spatio-temporal ones, it is not clear what the role of each feature is to the model's prediction or generation.
Designing explainable DL models for human mobility is essential to gain knowledge that can be useful for possible users, such as policymakers and urban planners.

\paragraph{Privacy.} DL models raise privacy issues both in the training and the prediction or generation phase. 
For example, in trajectory generation, evaluating the risk of re-identifying a real user from synthetic trajectories is crucial, especially when there is a scarcity of data to train the models.
A synthetic trajectory may resemble a real one and a malicious adversary may use this information for re-identification. 
In the training phase, the risk of leaking private data is high regardless of the mobility task, as the portions of information used cannot be controlled directly.
The extent training trajectories or crowd flows can be perturbed without degrading the realism of the generative models and the accuracy of predictive ones is an aspect that is barely investigated in the literature.

\paragraph{Tunability.}
Meaningful predictions and simulations require models that can be controlled along relevant mobility dimensions, such as geographic space, time granularity, presence of mobility restrictions or other events that may alter mobility, predictability of trajectories or crowd flows, and more.
Current DL models have a limited degree of tunability, which limits their usability in practice.
For example, it is not clear to what extent the models' realism or accuracy depends on the size or shape of the spatial tessellation, which may vary according to the user's needs.
For example, suppose the decision-maker is interested in identifying the most active areas of a city in terms of daily mobility. In that case, the usage of administrative tiles may provide more interpretable results.
In contrast, if they want to understand how the transportation and road networks affect and are affected by big events in a city, the usage of fine-grained grid-like tessellations may be helpful.

\paragraph{Interaction Dimension.}
Next-location predictors and trajectory generators assume the independence of the individuals' mobility, even though social purposes or collective needs can explain a significant portion of human movements.
People's movements are not entirely independent of each other and they can lead to situations like traffic jams, accidents, or massive commuting patterns.
Models that can include these aspects need to be designed to capture human mobility's complexity more comprehensively.

\begin{acks}
Luca Pappalardo has been partially supported by EU H2020 SoBigData++ grant agreement \#871042.
\end{acks}

\bibliographystyle{ACM-Reference-Format}
\bibliography{sample-base}

\newpage
\appendix

\section{Spatial tessellations}
\label{app:tessellations}
\begin{figure}[!htb]
	\centering
    \includegraphics[width=1\textwidth]{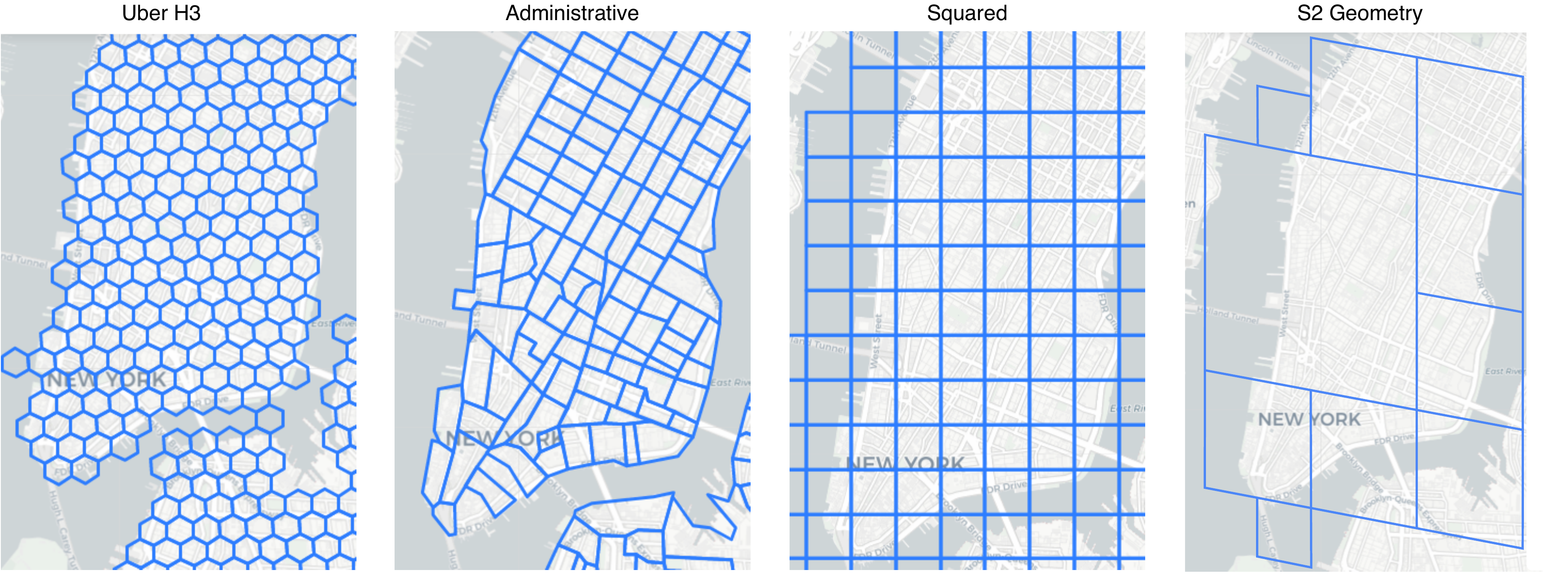}
\caption{Examples of spatial tessellations constructed over New York City. 
On the left, the city is tessellated in hexagons using H3 (resolution 6).
In the second image, the area is split according to the administrative boundaries described in the GeoJSON file downloaded from New York City's open data portal. 
In the third image, we use scikit-mobility to split an area into a squared grid with tiles of 1km$\times$1km.
In the last image, we make a tessellation using S2 Geometry.}
	\label{fig:grid}
\end{figure}

Spatial tessellations may be computed using several open-source tools.
Figure \ref{fig:grid} shows examples of tessellations in New York City.

Library scikit-mobility \cite{pappalardo2019scikit} allows for creating a squared spatial tessellation, given a reference geographic area and a tile size expressed in meters.

S2 Geometry is an open-source project that represents spatial data on a three-dimensional sphere.
It provides efficient and scalable spatial indexing techniques to carry out operations such as testing relationships among objects, measuring centroids, distances, and more.
S2 Geometry decomposes the unit sphere into a hierarchy of cells (tiles), each of which is a quadrilateral bounded by four geodesics. 
The top level of the hierarchy is obtained by projecting the six faces of a cube into the unit sphere, and lower levels are obtained by subdividing each cell into four sub-cells recursively.
Each cell in the hierarchy has a level, defined as the number of times the cell has been subdivided (starting with a face cell). Cells' levels range from 0 to 30. 
The smallest cells at level 30 are called leaf cells; there are $6 \times 4^{30}$ cells in total, each about $1 $cm across on the Earth's surface.

The H3 geospatial indexing system consists of a hexagonal tiling of the sphere with hierarchical indexes. 
The hexagonal grid system is created on the planar faces of a sphere-circumscribed icosahedron, and the grid cells are then projected to the surface of the sphere using a specific projection.
The H3 grid is constructed by recursively creating increasingly higher precision hexagon grids until the desired resolution is achieved. 
The first H3 resolution (resolution 0) consists of 122 base cells, and each subsequent resolution is created splitting each cell into seven children recursively.
H3 provides 15 finer grid resolutions in addition to resolution 0. 
The finest resolution, resolution 15, has cells with an area of less than $1m^2$.

\section{Deep Learning Modules}
\label{app:dl}
\subsection{Recurrent Neural Networks}
\label{app:rnns}

A fully connected neural network (FC) consists of a series of fully-connected layers. All the neurons in one layer are connected to the neurons in the next layer.
An FC layer, therefore, is a function that, given an input $x \in \mathcal{R}^m$, map values in $\mathcal{R}^m \rightarrow \mathcal{R}^n$ where $m$ and $n$ are the number of neurons in two consecutive layers.
FCs are universal approximators (i.e., can learn any representation function) \cite{goodfellow2016deep} 
but they are not able to deal with sequential data.

\begin{figure}[!htb]
	\centering
    \includegraphics[width=1\textwidth]{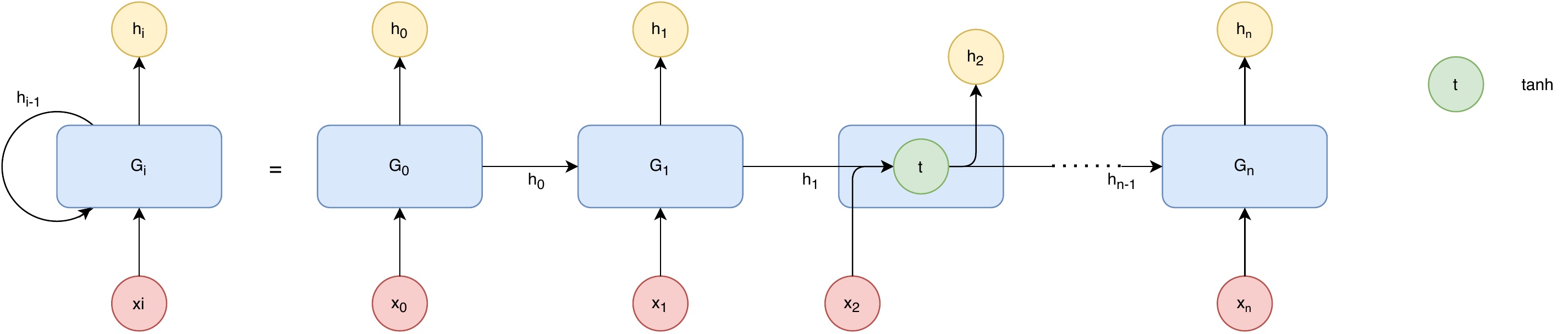}
    \hspace{2pt}
    \includegraphics[width=0.85\textwidth]{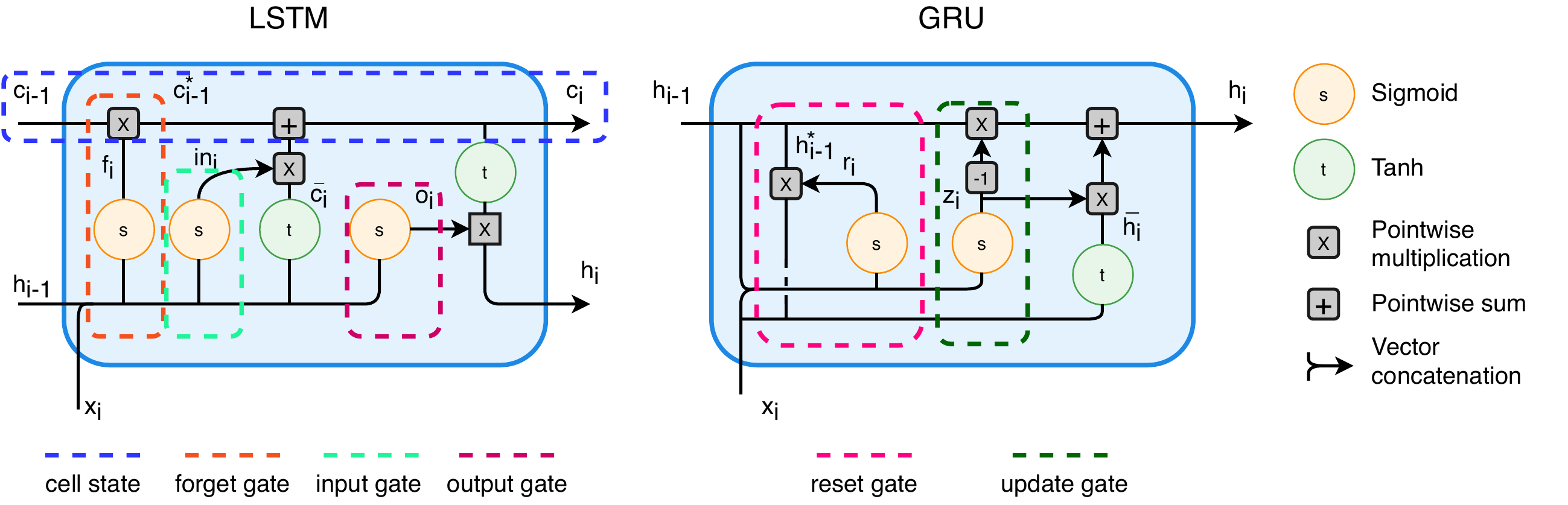}
\caption{
(Top) An example of a Recurrent Neural Network (RNN). 
The right term of the equation corresponds to the unrolled version of the network on the left term. 
In general, an RNN takes as input a sequence $X = \langle x_0, x_1, \dots x_n \rangle$ and produces an output $h_i$ that, at a certain moment $i$, is based on the current input $x_i$ and the output of the previous gate $h_{i-1}$.
Usually, in Vanilla RNNs, the tanh function (green circle) is used to combine the current input $x_i$ and the previous hidden state $h_{i-1}$.
(Bottom) The structure of an LSTM gate (left) and a GRU gate (right). 
In the LSTM gate, we highlight with a dashed line the cell state (in blue), the forget gate (in orange), the input gate (in green), and the output gate (in purple).
In the GRU gate, we highlight with the dashed line the reset gate (in red) and the update gate (in green).
}
	\label{fig:rnn_unfold}
\end{figure}

Recurrent Neural Networks (RNNs) \cite{10.5555/104279.104293} can efficiently deal with sequential data such as time series, in which values are ordered by time, or sentences in natural language, in which the order of the words is crucial to shaping its meaning. 
An RNN consists of a sequence of gates $G = \{ G_0, \dots, G_{n-1} \} $, each one producing an hidden state $h_i$ based on the current input $x_i$ and the output from the previous gate $h_{i-1}$ (Figure \ref{fig:rnn_unfold}).
In Vanilla RNNs, a gate $G_i$ is implemented by using a hyperbolic tangent function (tanh), which takes as input $x_i$ and $h_{i-1}$ and computes the current state $h_i$.
RNNs suffer from the vanishing gradient problem \cite{kolen2001gradient} and cannot propagate information found at early steps, losing relevant information at the beginning of a sequence when it is time to analyze its end \cite{kolen2001gradient}. 
Long-short-term Memory networks (LSTMs) \cite{hochreiter1997long} and Gated Recurrent Units (GRUs) \cite{cho-etal-2014-learning} are two gate implementations that mitigate this problem (Figure \ref{fig:rnn_unfold}).

In LSTMs, the cell state $c_i$ carries the data through the network, while the internal gates remove useless information from the flow or add relevant knowledge to the cell state.
There are three internal gates: forget, input, and output gate (Figure \ref{fig:rnn_unfold}). 
The forget gate passes the information of the previous hidden state $h_{i-1}$ and of the current input $x_i$ to a sigmoid activation function which ranges in $[0, 1]$. 
The closer the sigmoid's output ($f_i$) is to 0, the less likely the information is to be considered relevant. 
The sigmoid's output $f_i$ is then pointwise multiplied with previous cell state $c_{i-1}$ forming $c_{i-1}^*$. Values of $c_{i-1}^*$ close to 0 are not taken into consideration by the current cell state.
The input gate passes $h_{i-1}$ and $x_i$ to a sigmoid activation function and to a tanh activation function. 
The sigmoid function outputs $in_i$, i.e., the relevance of the new data (0 is not relevant; 1 is relevant).
The tanh outputs, $ \overline{c}_i \in [-1, 1]$, is multiplied with $in_i$, and the resulting product is pointwise summed with $c_{i-1}^*$ to obtain the new cell state $c_i$. 
The output gate generates the hidden state $h_i$. 
The previous hidden state $h_{i-1}$ and the current input $x_i$ are passed to a sigmoid activation function, which generates $o_i$. 
This value is pointwise multiplied with the output of a tanh applied on the new cell state $c_i$, to form the new hidden state $h_i$. 
The hidden state $h_i$ and the cell state $c_i$ are passed to the next gate of the recurrent network.
In some works, bidirectional LSTMs (Bi-LSTMs) \cite{650093} are adopted instead of LSTMs. 
Bi-LSTMs duplicate the recurrent layer: the sequence is provided as input to one layer and its reverse as input to the other layer. 
Analyzing time dependencies in both directions gives an advantage in scenarios like speech recognition, in which the context is essential to interpret the meaning of a sentence \cite{graves2005framewise, graves2013hybrid}.

In GRUs, the relevant information is propagated throughout the network using hidden states only (Figure \ref{fig:rnn_unfold}). 
GRUs have two types of internal gates.
The reset gate decides how much to forget and consists of a sigmoid function that takes as input the hidden layer of the previous step $h_{i-1}$ and the current input $x_i$.
The output of the sigmoid, $r_i$, is pairwise multiplied with $h_{i-1}$, generating $h_{i-1}^*$. 
Depending on $h_{i-1}^*$, we can determine which past information to forget (low values) or keep (high values).
The update gate establishes whether the past information is relevant for future predictions and, therefore, should be propagated to the next steps.
It is composed of a sigmoid function that takes as input $h_{i-1}$ and $x_i$ and outputs $z_i$. 
The next hidden state, $h_i$, is obtained by: \emph{(i)} computing $\hat{h}_i$, which is the output of a tanh function of $h_{i-1}^*$ and the current input $x_i$; \emph{(ii)} multiplying $1 - z_i$ with $h_{i-1}$; and \emph{(iii)} adding the resulting sum to the product between $z_i$ and $\hat{h}_i$.

Further mathematical details and formalizations can be found in \cite{sherstinsky2020fundamentals} for RNNs and LSTMs and \cite{cho2014learning} for GRUs.

RNNs are widely used in next-location prediction (Section \ref{sec:next}), often in combination with attention mechanisms, to capture the temporal relationships in individual trajectories.
Moreover, RNNs are often combined with convolutional neural networks in crowd flow prediction (Section \ref{sec:crowd}) and trajectory generation (Section \ref{sec:task_generative}), to capture temporal and spatial patterns at the same time.

\subsection{Convolutional Neural Networks}
\label{app:cnns}

Convolutional Neural Networks (CNNs) are widely used in computer vision for their efficacy in object recognition \cite{simonyan2014very, ren2015fast}, image classification and segmentation \cite{krizehvsky2012imagenet, farabet2012learning}, movement or event recognition \cite{toshev2014deeppose}, and more \cite{ khan2020survey}. Similarly to the visual cortex \cite{hubel1959receptive, hubel1962receptive}, a CNN is a network made of neurons that react only to certain stimuli in a restricted region of the visual field \cite{krizehvsky2012imagenet}. 

\begin{figure}[!htb]
	\centering
    \includegraphics[width=1\textwidth]{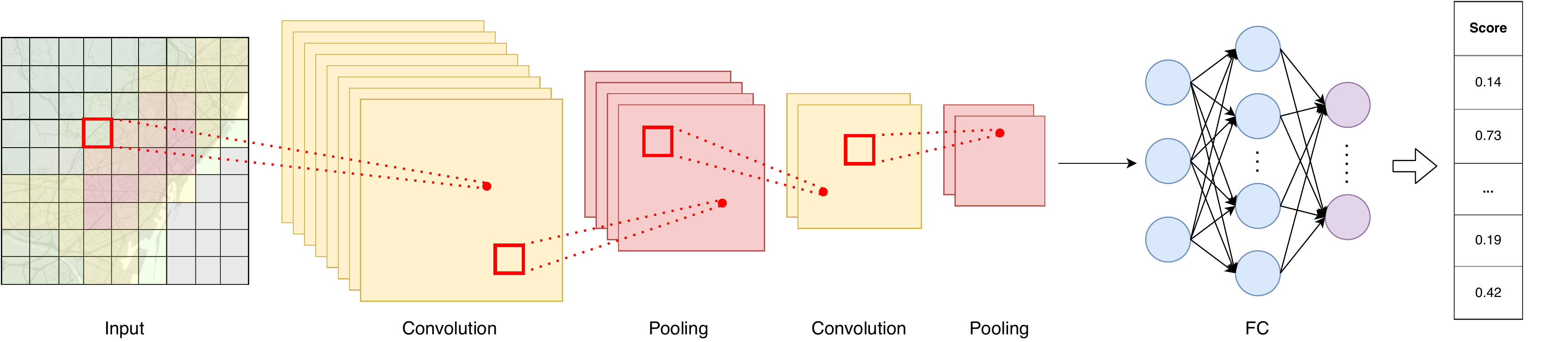}
\caption{Example of architecture of a CNN. 
The neural network consists of two convolutions and two pooling layers. 
After the convolution/pooling layers, there is an FC that outputs the prediction. 
For instance, to classify a sample into one of $n$ classes, the last layer is usually fed into an FC that outputs a $1 \times n$ vector containing the probabilities of the sample to belong to a specific class.}
\label{fig:struct_cnn}
\end{figure}

CNNs alternate two types of layers: \emph{(i)} the convolutional layers reduce the size of the matrix by applying a kernel function, or filter, that keeps all the relevant information; \emph{(ii)} the pooling layers reduce the spatial size of the convoluted features to decrease the computational power required to process the data. 
Figure \ref{fig:struct_cnn} shows an example of a CNN architecture with two convolutional layers and two pooling layers. 
Usually, at the end of the CNN, an FC is used to compute the output. 

The convolutional layers apply one or more filters to the input matrix $A$ to extract relevant features and summarize the characteristics of
an $i \times j$ area of $A$ into a single value. 
Specifically, given an input matrix $A$ of size $n \times m$, a filter is a mathematical operation between the original matrix $A$ and another matrix $B$ of size $k \times l$, with $k < n$ and $l < m$.
The filter is generally applied with a sliding window mechanism called stride. 

The pooling layers aim at downsampling the input matrix, replacing each portion of it with summary statistics. 
In practice, pooling layers are either max pooling or average pooling. 
Max pooling returns the maximum value from the part of the matrix convoluted by a given size and stride filter. 
Average pooling returns the average of the values produced by the filter operations. 
Pooling layers create representations of the matrix that are approximately invariant to small translations of the input.

A significant limitation of CNNs, especially relevant for networks with many layers, is the vanishing gradient issue \cite{kolen2001gradient}. The so-called residual units are a solution to this problem \cite{he2016deep}. 
A residual unit implements the skip connection also known as identity connection. Given a network with $l_1,\dots, l_k$ layers, the output of a layer $l_i$ is added to the output of the layer $l_{i+j}$ so to preserve the loss of information. The variable $j$ represents the skip size, and it is usually smaller than 4.

A mathematical formalization of CNNs can be found in \cite{wu2017introduction}.

Differently from RNNs, where the input is processed sequentially, CNNs are not designed to process sequences, and they are used mainly in crowd flow prediction (Section \ref{sec:crowd}): the evolution of the incoming (outgoing) flows within a city is represented as a sequence of matrices. CNNs are used to capture the dynamics of the spatial dependencies among the areas of a city. 
CNNs are often combined with RNNs to capture temporal dependencies (e.g., ConvLSTM \cite{xingjian2015convolutional}).
CNNs are also used in trajectory generation (Section \ref{sec:task_generative}): real trajectories are represented as images and used to train a Generative Adversarial Network (GAN), in which two CNNs are used as generator and discriminator to generate realistic trajectories.

\subsection{Attention Mechanism}
\label{app:attention}
Attention mechanisms are based on the idea that, when dealing with a large amount of information, our brains focus on the most significant parts and consider all the others as background information. 
Initially introduced for natural language processing (e.g., machine translation \cite{cho2014learning, sutskever2014sequence} and speech recognition tasks \cite{chorowski2015attention}), the usage of attention mechanisms rapidly extended to computer vision, healthcare, and recommendation systems \cite{xu2015show, seo2017interpretable, choi2016retain}.
Given the current input $x$ and a context, an attention mechanism produces a score for each element of $x$. 
These scores are usually computed using adequate activation functions (e.g., softmax) and organized in a vector $s$. 
A context vector $\hat{c}$ is computed as the pairwise multiplication between $s$ and $x$. 
The context plays a crucial role in the establishment of which features should be the most important to the model. 

\subsection{Generative Models}
\label{app:generative}

Dimensionality reduction consists in transforming data from a high-dimensional space to a low-dimensional one retaining important properties and information of the original data \cite{kramer1991nonlinear}. 
In general, this is done by employing a module called \textit{encoder} (i.e., a neural network), which reduces the space of the original features generating a \emph{latent space}; and a \textit{decoder} (i.e., a neural network), which transforms the latent space back to the original space. 
AutoEncoders (AEs) \cite{kramer1991nonlinear} reduce the reconstruction error while decoding. AEs are designed to encode a sample always in the same way and, therefore, cannot be considered a generative model as they are not capable of decoding (i.e., generate) data with different charateristics with repsect to the original data. Variational AutoEncoders (VAEs) \cite{kingma2013auto} solve this problem by encoding a sample as a distribution instead of a set of fixed values. In this way, the decoder can be used to generate data that are similar to the original ones (e.g., realistic as they should follow similar patterns) but diverse.

Generative adversarial networks (GANs) \cite{goodfellow2014generative} are an example of generative models, i.e., any model that takes a training set, consisting of samples drawn from a distribution, and learns to represent an estimate of that distribution.  
A generative mobility model can generate synthetic spatio-temporal trajectories that realistically reproduce mobility patterns. 
Recently, GANs are being used to generate synthetic mobility trajectories \cite{liu2018trajgans, ouyang2018non}.

\begin{figure}
    \centering
    \includegraphics[width=0.8\textwidth]{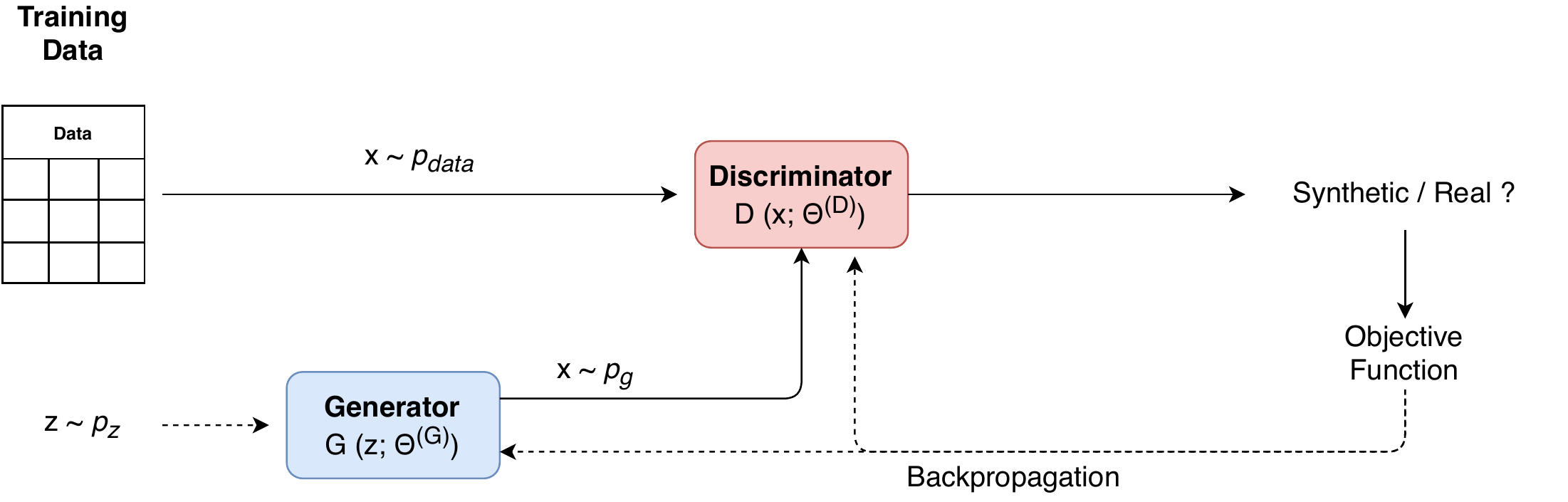}
    \caption{
    Visual representation of a Generative Adversarial Network (GAN). 
    A GAN is composed of a generator $G$, and a discriminator $D$. 
    The generator is a differentiable function $G(z; \Theta^{(G)})$ which outputs the new data according to a distribution $p_g$, where $\Theta^{(G)}$ are the parameters of the generative model. 
    The discriminator represents a differentiable function $D(x; \Theta^{(D)})$, where $\Theta^{(D)}$ are the parameters of the discriminative model, which produces the probability that $x$ comes from the distribution of training data $p_{data}$. 
    The aim is to obtain a generator that is a good estimator of $p_{data}$. 
    When this occurs, the discriminator is "fooled" and can no longer distinguish the samples from $p_{data}$ from those from $p_g$.
    }
    \label{fig:gans}
\end{figure}

The basic idea of a GAN is to set up a game between a generator (e.g., a neural network) and a discriminator (e.g. a classifier) (Figure \ref{fig:gans}). 
The generator creates samples from the distribution $p_g$ that are intended to come from the same distribution as the training data $p_{data}$ and hence similar to the original ones.
The discriminator examines the generated samples to determine whether they are real or fake. In other words, the generator is trained to fool the discriminator by generating samples that are indistinguishable from real ones.

Mathematically, the generator and the discriminator are two functions that are differentiable with respect to both the inputs and the parameters. 
The discriminator is a function $D$ that takes $x$ (e.g., a trajectory) as input and uses $\Theta^{(D)}$ as parameters. 
The generator is a function $G$ that takes $z$ (e.g., set of trajectories) as input and uses $\Theta^{(G)}$ as parameters.
$D$ wishes to minimize a cost function $J(D)(\Theta^{(D)}, \Theta^{(G)})$ while controlling only $\Theta^{(D)}$. 
Similarly, $G$ wishes to minimize a cost function $J(G)(\Theta^{(D)}, \Theta^{(G)})$ while controlling only $\Theta^{(G)}$. 
This scenario is a game the solution of which is a Nash equilibrium \cite{goodfellow2016deep}, i.e., a tuple $(\Theta^{(D)}, \Theta^{(G)})$ that is a local minimum of $J(D)$ with respect to $\Theta^{(D)}$ and a local minimum of $J(G)$ with respect to $\Theta^{(G)}$.

The training process consists of two simultaneous Stochastic Gradient Descent (SGD) \cite{goodfellow2016deep}. 
SGD is a stochastic approximation of gradient descent optimization that replaces the actual gradient with an estimate, thus achieving faster iterations in trade for a lower convergence rate.
Two gradient steps are made simultaneously: one updating $\Theta^{(D)}$ to reduce $J(D)$ and one updating $\Theta^{(G)}$ to reduce $J(G)$. 
Usually, the cost function used for the discriminator and the generator is the cross-entropy \cite{goodfellow2016deep}, defined as:
$ H(P,Q) = -E_P [ log Q]$,
where $P, Q$ are two distributions and $E_P$ is the expected value of $log Q$ according to $P$. 

A formalization of the concept of GAN can be found in \cite{goodfellow2014generative}

\section{Data}
\label{app:data}
The last decade has witnessed the emergence of massive datasets of digital traces that portray human movements at an unprecedented scale and detail. 
Examples include tracks generated by GPS devices embedded in smartphones \cite{zheng2010geolife} or private vehicles \cite{pappalardo2013understanding}; 
mobile phone records  \cite{blondel2015survey}; and geotagged posts from social media platforms \cite{noulas2012tale}. 
Unfortunately, most of these datasets are proprietary and not publicly available, making research on human mobility hard to reproduce.
{\color{black}In this Section, we discuss the peculiarities of various mobility data sources and, for each of them, provide a reference to a list of public datasets commonly used to train and test models presented in this survey (Table \ref{tab:open_datasets}).}

\subsection{Mobile Phone Records} 
\label{sec:mobile_phone_data}
Mobile phones are ubiquitous, with coverage in most countries that reaches almost 100\% of the population \cite{blondel2015survey, itu2019}. 
Telco companies record the activity of mobile phone users for billing and operational purposes, hence storing an enormous amount of information on where, when, and with whom users communicate \cite{blondel2015survey}. 
Every time a user engages a telecommunication interaction -- calls, text messages, data connections -- the operator assigns a Radio Base Station (RBS) to deliver the communication through the network. 
Since the position and the coverage area of each RBS are known, a user's telecommunication interaction reflects their geographic location.
Each interaction generates several mobile data formats \cite{blondel2015survey, pappalardo2020individuallevel}. 

Every time a user makes/receives a call or sends/receives an SMS, a new Call Detail Record (CDR) is created. 
A CDR is a tuple $(\mbox{u}_o, \mbox{u}_i, t, A_o, A_i, d)$, where $\mbox{u}_o$ and $\mbox{u}_i$ are the identifiers of the caller and the callee, and $t$ is a timestamp of when the call starts. 
In turn, $A_o$ and $A_i$ are the RBSs that manage the outgoing call and the receiving call, and $d$ is the call duration.  
An individual's mobility can be reconstructed from CDRs assuming a movement between the RBSs of any two consecutive records \cite{blondel2015survey}. 
Aggregated mobility, such as flows, can be inferred by counting the number of users that move, in a given time window, between two RBSs or spatial aggregations of them (e.g., neighborhoods or municipalities) \cite{calabrese2011estimating}.
CDRs are sparse in time, i.e., a user's position is known when they make or receive a call or a text message only, leading to sparse and incomplete mobility trajectories.  
Notwithstanding, they are the most common format in human mobility studies \cite{gonzalez2008understanding, simini2012universal, csaji2013exploring, barbosa2015effect, pappalardo2015returners, blondel2015survey}.

When a user uploads or downloads data from the Internet using their phone's connection, they generate an eXtended Detail Record (XDR), a tuple $(\mbox{u}, t, A, k)$ where $A$ is the RBS that serves the connection, and $k$ the amount of uploaded/downloaded information. XDRs are less common in the literature than CDRs, mainly because the advent of the mobile data connection is relatively recent. 
XDRs partially overcome the problem of sparsity present in CDRs \cite{chen2019complete}, because mobile connections are more frequent than calls and text messages.

For both CDRs and XDRs, the spatial granularity is at the level of RBS, i.e., the user position is approximated with the location of the RBS used for the telecommunication activity. 
This approximation implies that the user's position within an RBS's coverage area is unknown and that user tracking depends on the spatial distribution of antennas on the territory, which depends on population density. 
Nevertheless, mobile phone data may cover a large sample size on a national scale. 
An advantage of mobile phone data is its multidimensionality: CDRs also provide information about the social interactions between users; both CDRs and XDRs may be accompanied by socio-demographic information about the users (e.g., age, sex).  

\subsubsection{Available datasets.}
Since mobile phone data contain sensitive information \cite{pellungrini2017data, demontoye2018privacy}, they are not typically publicly available, and any data collected by a specific group of researchers may not be shared with other groups, making reproducibility difficult.

\subsection{GPS traces}
 \label{app:gps}

Global Navigation Satellite Systems (GNSS) use satellites to provide geo-spatial positioning, allowing electronic receivers to determine their location (longitude, latitude, and altitude) and time, using signals transmitted along a line of sight from satellites. 
The US Global Positioning System (GPS) is the most popular one,
and GPS receivers are ubiquitous in many tools of everyday life, such as mobile phones \cite{zheng2010geolife, stopczynski2014measuring, alessandretti2018evidence}, vehicles \cite{pappalardo2013understanding, pappalardo2015returners, gallotti2015stochastic}, vessels \cite{russo2016assessing, vespe2016mapping, Sleimani2015anomaly}, and wearable devices \cite{rossi2018effective}.
On mobile phones, GPS receivers are activated by apps that require the user's position (e.g., Google Maps).
On vehicles, GPS devices automatically turn on when the vehicle starts, sending positions to a server with a frequency of a few seconds. 
The precision of GPS receivers varies from a few centimeters to meters, depending on the device's quality and the errors generated by the system \cite{Carlson-2010-10506}.
GPS cannot track the devices in enclosed spaces, such as buildings and tunnels.
A typical GPS trace is a set of tuples $(u, t, lat, lng)$ where $u$ is a user, $t$ is a timestamp of the measurement and $lat, lng$ are the position's latitude and longitude. 
GPS traces may require several preprocessing tasks aimed to mitigate errors and extract meaningful semantics. 
For example, since GPS traces are dense sequences of spatio-temporal points, they do not explicitly define semantic locations, which must be inferred through specific preprocessing techniques \cite{feng2016survey,zheng2014urban, zheng2015trajectory}. 

\subsubsection{Available datasets}
GeoLife \cite{zheng2010geolife} is a publicly available dataset describing the GPS trajectories of 182 users over 4.5 years, collected using different devices (e.g., GPS receivers and mobile phones) every 1-5 seconds or 5-10 meters.
GeoLife covers a broad range of users' outdoor movements, from life routines like go home and go to work to entertainment and sports activities. 
Each point in a GeoLife's trajectory contains latitude, longitude, altitude, and the timestamp.
Moreover, the user's transportation mode is also available for most of the trajectories.

Other public datasets provide information about the trips of GPS-equipped taxis in several cities. 
Piorkowski et al. \cite{epfl-mobility-20090224} provide the trajectories, sampled on average every 10 seconds, of taxis in San Francisco, in May 2008. 
Each point of a trajectory consists of the taxi's identifier, the latitude, the longitude, the timestamp, and the occupancy. 
Moreira et al. \cite{moreira2013predicting} (ECML/PKDD Challenge) provide the trips of taxis in Porto, Portugal, in which points contain the latitude, the longitude and a timestamp indicating when the trip started. Data are collected approximately every 15 seconds.  
For each trip, the dataset provides several auxiliary information, such as the trip's typology (e.g., dispatched from the central, demanded to the operator, requested to the driver), the stand from which the taxi departed, and an identifier of the passenger's phone number. 
The Taxi and Limousine Commission of New York City collected a dataset on yellow and green taxis operating in the city starting from 2009 \cite{tlc}. The dataset provides information on pick-up and drop-off dates/times and locations, trip distances, itemized fares, rate types, payment types, and driver-reported passenger counts.

The T-Drive dataset \cite{zheng2011t-drive}  describes the trajectories of about 10,000 taxis in Beijing, China, for one week.
Points are sampled every 177 seconds and contain the taxi's identifier, the latitude, the longitude, and the timestamp. 
Zhang et al. \cite{zhang2017deep} create a squared tessellation on New York City and Beijing and, for each tile, provide the incoming and the outgoing flows. 
Beijing's flows are captured based on taxis' GPS signal; the New York City ones are based on the city's bike-sharing system. 

Similarly, Jiang et al. \cite{jiang2021deepcrowd} generated a tessellation of 450 $\times$ 450 meters over Tokyo and Osaka, Japan and, for each tile, they provide people density and incoming and outgoing flows. The mobility is captured using GPS devices.

Citi Bike by Lyft \cite{bikenyc} provides a dataset describing the trips between bike-sharing stations in New York City. 
Each record describes the stations where the trip started and ended when it took place and the stations' coordinates. 
There are similar open datasets for other cities, such as Washington DC \cite{bikewo}.

The Mobile Data Challenge dataset (MDC) \cite{laurila2012mobile} describes the trajectories of 185 participants of a data collection campaign in Lausanne, collected from mobile phones  \cite{kiukkonen2010towards}.
The dataset offers various information (e.g., calendar logs, data from accelerometers), including two files describing the individuals' mobility through GPS receivers and Wireless Local Area Networks (WLANs). 
Each record in the first file describes the user's identifier, latitude, longitude, altitude, timestamp, speed, heading direction, and accuracy. 
Records in the second file describe the user's identifier, a timestamp, the first three bytes of the device's MAC address, and the access points' location. 

Recently, to take countermeasures for the COVID-19 pandemic, companies such as Cuebiq and SafeGraph are providing free access upon request to their data.\footnote{Cuebiq: \href{https://www.cuebiq.com/visitation-insights-covid19/}{cuebiq.com/visitation-insights-covid19/}, SafeGraph: \href{https://www.safegraph.com/covid-19-data-consortium}{safegraph.com/covid-19-data-consortium}. For instance, \cite{kang2020multiscale} is a recently published dataset based on SafeGraph. It contains aggregated daily mobility flows at different spatial aggregations such as country to country, state to state and census tract to census tract flows.} 

\subsection{Social Media Data}
\label{app:social_media}

Users' posts on social media (e.g., photos, text, videos) can be associated with a geographic location and time (geotagged).
The presence of spatial and temporal information allows the reconstruction of users' trajectories from the sequence of published posts. 
Some platforms like Twitter provide either the post's precise geotag (i.e., a latitude and longitude pair, a format recently removed) or the position of a predefined location suggested by the platform (e.g., a city, an area, a restaurant). 
In other platforms like Foursquare and Facebook, users can check-in in predefined locations called venues, i.e., POIs that provide information about social, cultural, and infrastructural components of a geographic area (e.g., cities, shops, museums).
A venue is associated with a physical location (latitude and longitude pair) and textual information (a description of the place or the activities related to the place) and can follow a hierarchical categorization that provides different levels of detail about the activities (e.g., Food, Asian Restaurant, Chinese Restaurant) \cite{barlacchi2017structural}. 

In general, a geotagged record describes the posting user's identifier, the resource identifier (e.g., post, tweet, photo), the time of posting, and, depending on the platform, the venue identifier/category or a location as a string (e.g., Statue of Liberty, New York) or a latitude/longitude pair.
 
For most of the social media platforms, geotagged posts are downloadable through their APIs. 
APIs impose limitations on the number of downloadable posts and queries per day or require authorization from the platform's users to download the data.
Users' location is available only when they post something or check-in into a venue, leading to a data sparsity problem. 
Nevertheless, social media data bring the advantage that an objective definition of location is available, facilitating the data preprocessing phase \cite{cui2018social}.

\subsubsection{Available datasets}
Datasets about check-ins on social media platforms that are not active anymore, such as Gowalla and Brightkite, are freely available. 
Gowalla was a location-based social network platform in which, similarly to Foursquare, the users were allowed to check-in in the so-called spots (venues) through a website or the app. 
The related dataset \cite{cho2011friendship} consists of more than six million check-ins over one year and a half from February 2009 to October 2010. 
Each check-in describes the user identifier, the location identifier, the latitude and longitude pair, and a timestamp. 
The dataset also provides information about the users' friendship network, which contains about 200,000 nodes and one million edges \cite{cho2011friendship}. 
In Brightkite, another platform that is not active anymore, users were allowed to check-in in POIs, to specify who is nearby at the moment and who went to a POI before. 
The dataset contains almost 4.5 million check-ins from April 2008 to October 2010 and the users' friendship network with about 60,000 nodes and 220,000 edges \cite{cho2011friendship}. 

A dataset collected through Foursquare APIs is introduced in \cite{feng2018deepmove} and contains check-ins of 16,000 users over one year in New York City.

Twitter provides several open datasets, in which location is usually expressed as a semantic point of interest either suggested by the platform (e.g., Empire State Building, NYC) or typed by the users (e.g., Home), or as a latitude and longitude pair. 
Geotagged tweets may be retrieved directly using Twitter APIs.
For example, Zhang et al. \cite{zhang2016gmove} (GMove) provide a Twitter dataset describing 1.4 million tweets from August to November 2014 covering the area of Los Angeles. 

A comprehensive list of Twitter datasets is available at \href{https://github.com/shaypal5/awesome-twitter-data}{github.com/shaypal5/awesome-twitter-data}.

\begin{table}[]
    \scriptsize
    \begin{tabularx}{\textwidth}{c c X r c X X X X}
      \toprule
      & \scriptsize \textbf{Ref.}   &   \scriptsize \textbf{Name}       &   \scriptsize \textbf{Items} &   \scriptsize \textbf{Time span}   &   \scriptsize \textbf{Area}     &   \scriptsize \textbf{Used By} & \scriptsize \textbf{Task} &   \scriptsize \textbf{Link (https://bit.ly/)} \\ \midrule

      \multirow{14}{*}{\rotatebox[origin=c]{90}{\bf GPS traces}} & \cite{zheng2010geolife}          &   GeoLife                                                             &    182                             &   4.5 Years                         &   Asia                                           &   \cite{feng2020learning} & Traj. Gen. & \href{https://bit.ly/Geolife}{Geolife} \\ 
       
       & \cite{zheng2011t-drive}  &   T-Drive & 10k  &   1 Week  &   Beijing, China  &   \cite{iwata2019neural} & Flow. Pred. & \href{https://bit.ly/T-Drive-Data}{T-Drive-Data} \\ 
       
       & \cite{jiang2021deepcrowd} &   DeepCrowd & - & 4 months & Tokyo and Osaka, Japan & \cite{jiang2021deepcrowd} & Crowd Flow Pred. & \href{https://bit.ly/DeepCrowd}{DeepCrowd}\\
       
        & \cite{zhang2017deep} &   ST-ResNet taxis & - & 4, 6 months & Beijing, China & \cite{ren2020hybrid, yuan2020deep, liu20120dynamic, li2019densely, sun2019predicting, du2019deep, jin2018st, zonoozi2018periodic, zhang2017deep} & Crowd Flow Pred. & \href{https://bit.ly/ST-ResNet}{ST-ResNet}\\
       
       & \cite{zhang2017deep} &   ST-ResNet bikes & - & 6 months & New York City, USA & \cite{ren2020hybrid, yuan2020deep, liu20120dynamic, li2019densely, sun2019predicting, du2019deep, jin2018st, zonoozi2018periodic, zhang2017deep} & Crowd Flow Pred. & \href{https://bit.ly/ST-ResNet}{ST-ResNet}\\
    
       & \cite{epfl-mobility-20090224} & Taxi San Francisco  & 500 & 30 days & San Francisco, USA  & \cite{ebel2020destination, rossi2019modelling, yin2018gans} & Next-Loc., Traj. Gen. & \href{https://bit.ly/TaxiSF}{TaxiSF} \\ 

       & \cite{moreira2013predicting}                          &   ECML-PKDD taxi  & 441 & 9 months & Porto, Portugal  & \cite{ebel2020destination, rossi2019modelling, lv2018t, de2015artificial} & Next-Loc. & \href{https://bit.ly/TaxiPorto}{TaxiPorto} \\ %
       
       & \cite{tlc} & Taxi New York City & - & From 2009 & New York City & \cite{rossi2019modelling, tian2020deep, sun2019predicting} & Next-Loc. & \href{https://bit.ly/TaxiNYC-2}{TaxiNYC-2} \\

       & \cite{laurila2012mobile} & MDC & 185 & 2 years &   Lausanne, Switzerland & \cite{ouyang2016deepspace, kulkarni2018generative} & Traj. Gen & \href{https://bit.ly/MDC-2}{MDC-2} \\

        & \cite{kang2020multiscale}   & COVID 19 US Flows  & - & From 2019  & United States & \cite{simini2020deep} & Flow Gen. & \href{https://bit.ly/USFlows}{USFlows} \\
    
      \midrule
    
      \multirow{8}{*}{\rotatebox[origin=c]{90}{\bf check-ins}} &  \cite{cho2011friendship}         &   Gowalla &    196k   &   20 months  &   California \& Nevada, USA   & \cite{yang2020location, gao2019predicting, liu2016predicting} & Next-Loc. & \href{https://bit.ly/GowallaData}{GowallaData} \\ %
        
      &  \cite{cho2011friendship}         &   Brightkite   &  58k  &   30 months  & - & \cite{yang2020location, gao2019predicting, liu2016predicting} & Next-Loc. & \href{https://bit.ly/Brightkite}{Brightkite}  \\ %
         
      &    \cite{feng2018deepmove}       &   DeepMove    &   16k     & 1 Year      &  New York City  & \cite{feng2018deepmove}  &  Next-Loc. & \href{https://bit.ly/DeepMove}{DeepMove} \\ %

     &    \cite{zhang2016gmove}       &   GMove    &    1.4M    &   4 Months    & Los Angeles  & \cite{yao2017serm} & Next-Loc. & \href{https://bit.ly/SERM-Repo}{SERM-Repo} \\ %
     

      & \cite{bikenyc}                   &   New York City bikes   &  -  & from 2013 &   New York City, USA &   \cite{tian2020deep, sun2019predicting, lin2019deepstn, yao2018modeling} & Crowd Flow Pred.& \href{https://bit.ly/BikeNYCData}{BikeNYCData} \\

      & \cite{bikewo}  &   Washington DC bikes  & - & from 2010 &   Washington DC , USA  & \cite{sun2019predicting} & Crowd Flow Pred. & \href{https://bit.ly/BikeWashington}{BikeWashington} \\

       \bottomrule

    \end{tabularx} 
    \caption{Public mobility datasets used in the selected papers. 
    For each dataset, we provide a reference to the paper introducing it, the number of items (users or points) in the dataset (symbol ``-'' indicates that the dataset is aggregated, that the number is not available, or that the dataset is continuously updated), its time span, the geographic area covered, the list of selected papers that use it, the mobility tasks the dataset is used for, and the link to download it.}
     \label{tab:open_datasets}
\end{table}

\section{Evaluation Metrics}
\label{app:evaluation_metrics}

\subsection{Distance metrics}
\label{app:distance_metrics}
The \textit{Haversine distance} is the distance on the spherical earth\footnote{Flat earthers can simply use the Euclidean distance, defined as follows: given two points $p_1 = (x_1, y_1)$ and $p_2 = (x_2, y_2)$, $d_{\text{flatearth}} {=} \sqrt{(x_1 - x_2) ^ 2 + (y_1 - y_2) ^ 2}$} between two points $p_1$ and $p_2$:
\begin{equation}
\small
d_{h}(p_1,p_2)=2R\left(\sqrt{\frac{a(p_1,p_2)}{a(p_1,p_2)-1}}\right); \mbox{\quad}  a(p_1,p_2)=sin^2\left(\frac{\phi_2-\phi_1}{2}\right)\\ +cos(\phi_1)cos(\phi_2)sin^2\left(\frac{\lambda_2-\lambda_1}{2}\right)
\end{equation}
where $R$ is the earth radius and $\lambda_i$ and $\phi_i$, with $i=1,2$, are the longitude and the latitude of $p_i$, respectively. The Haversine distance range in $[0, \infty]$ and lower values indicate better performance.
The \textit{equirectangular distance} is defined as:

\begin{equation}
\small
    d_{eq}(p_1, p_2) = R \sqrt{\left ((\lambda_{p_2} - \lambda_{p_1}) cos \left (\frac{\phi_{p_2} - \phi_{p_1}}{2} \right ) \right) ^2 + (\phi_{p_2} - \phi_{p_1})^2}
\end{equation}
where $\lambda_{p_i}$ and $\phi_{p_i}$ are the longitude and latitude of point $p_i$, respectively.

\subsection{Classification metrics}
\label{app:classification_metrics}
Accuracy (ACC) indicates how many of the locations an individual will visit are correctly predicted.
The k-accuracy (ACC@k) is often used instead of ACC: predictors output a list of all possible locations an individual can visit next, ranked from the most to the least likely, and ACC@k is the fraction of times the real location is among the $k$ most likely locations predicted by the model, i.e., the percentage that a list of predictions with length $k$ covers the ground truth location (ACC = ACC@1).

Precision measures how accurate the predictor is on the positive class. 
Recall measures the True Positive Rate (TPR), i.e., the fraction of positive instances correctly predicted by the model. 
F1-score summarizes the performance of a model and it is computed as the harmonic mean of Precision and Recall. 
Given the number of True Positives (TPs), False Positives (FPs), and False Negatives (FNs), we define 
Precision, Recall, and F1-Score as:
 
\begin{equation}
\small
Precision = \frac{TP}{(TP+FP)}; \mbox{\quad} Recall = \frac{TP}{(TP+FN)}; \mbox{\quad} F1 = 2 \frac{Precision \times Recall}{ Precision + Recall}
\end{equation}

The Receiver Operating Characteristic Curve (ROC) visualizes a classifier's performance by plotting the  Recall against the False Positive Rate (FPR). 
FPR is the ratio of FPs over the sum of FPs and True Negatives (TNs). 
AUC ($\in [0, 1]$) measures the area under the ROC. 
It is scale-invariant and provides an aggregate measure of performance across all possible classification thresholds.
The higher the AUC, the better the model, where AUC=0.50 indicates the performance of a model that makes predictions at random. 

\subsection{Error metrics}
\label{app:error_metrics}
Commonly used error metrics are Mean Average Error (MAE), Mean Squared Error (MSE), Root Mean Squared Error (RMSE), and MAPE (Mean Average Percentage Error), defined as follows:
\begin{equation}
\small
\mbox{MAE} = \frac{1}{n} \sum_{i=1}^n |y_i - \hat{y}_i|; 
\mbox{\quad}
\mbox{MSE} = \frac{1}{n} \sum_{i=1}^n (y_i - \hat{y}_i)^2;
\mbox{\quad}
\mbox{RMSE} = \sqrt{\frac{1}{n} \sum_{i=1}^n (y_i - \hat{y}_i)^2 };
\mbox{\quad} \mbox{MAPE} = \left(\frac{1}{n} \sum_{i=1}^n \frac {|y_i - \hat{y}_i|}{|y_i|} \right) * 100 
\label{eq:mae}
\end{equation}
where $\hat{y}_i$ indicates the predicted value, $y_i$ indicates the actual value, and $n$ is the number of predictions.
All metrics range in $[0, \infty]$ and lower values indicate better performance.
Since MAE uses the error's absolute value, it does not consider whether the model overestimates or underestimates the actual value. 
MSE weighs large errors more than MAE, and it is sensitive to outliers.
RMSE weighs the errors more than MAE, hence penalizing models that produce large errors. 
As the values are squared, the RMSE is expressed in the same unit as the predicted one.
 
The S{\o}rensen-Dice index, also called Common Part of Commuters (CPC) \cite{barbosa2018human,lenormand2016systematic}. 
It is a well-established measure to compute the similarity between real flows, $y^r$, and generated flows, $y^g$: 
\begin{equation}
    CPC = \frac{2 \sum_{i,j} min (y^g(l_i, l_j), y^r(l_i, l_j))}
    {\sum_{i,j} y^g(l_i, l_j) + \sum_{i,j} y^r(l_i, l_j)} 
\end{equation}
CPC is always positive and contained in the closed interval $(0, 1)$ with 1 indicating a perfect match between the generated flows and the ground truth and 0 highlighting bad performance with no overlap. 
 
\subsection{Divergence metrics}
\label{app:divergence}
The Kullback-Leibler (KL) divergence measures how different a probability distribution is from a reference probability distribution.
It is used to assess the performance of a generative mobility model by calculating the extent to which synthetic trajectories and real trajectories are similar with respect to relevant mobility patterns.
Formally, given two discrete probability distributions $P$ and $Q$, defined on the same probability space $X$, the KL divergence from $P$ to $Q$ is defined as:
\begin{equation}
\small
D_{\mbox{\scriptsize KL}} (P||Q) = \sum_{x \in X} P(x) log \left( \frac{P(x)}{Q(x)} \right).
\end{equation}

Formally, given two probability distributions $P$ and $Q$,  KL divergence is the expectation of the logarithmic difference between the probabilities of $P$ and $Q$, where the expectation is taken using the probabilities of $P$.
KL divergence is always non-negative ($D_{\mbox{\scriptsize KL}}(P||Q) \geq 0$) and not symmetric, i.e., $D_{\mbox{\scriptsize KL}}(P||Q) \neq D_{\mbox{\scriptsize KL}}(Q || P)$. 
$P$ and $Q$ are the same distribution if $D_{\mbox{\scriptsize KL}}(P||Q) = 0$.

The Jensen-Shannon (JS) divergence is a measure to assess the similarity between two distributions.
It is based on the KL divergence but it is symmetric ($JS ( P || Q) = JS (Q || P)$) and ranges in $[0, 1]$. 
Formally, given two probability distributions $P$ and $Q$, and $M = \frac{1}{2}(P||Q)$, we define the JS divergence as:
\begin{equation}
\small
D_{\mbox{\scriptsize JS}} (P || Q) = \frac{1}{2} D_{\mbox{\scriptsize KL}}(P || M) + \frac{1}{2} D_{KL}(Q || M).
\end{equation}
The JS divergence is used, as an alternative to KL, to assess the performance of generative mobility models.

\section{Human Mobility Patterns}
\label{app:mobility_patterns}
Human movements, far from being random, follows well-defined statistical patterns \cite{barbosa2018human, wang2019urban}. 
In this Section, we revise the most relevant spatial (Section \ref{sec:spatial_metrics}) and temporal (Section \ref{sec:temporal_metrics}) patterns of human mobility.

\subsection{Spatial patterns}
\label{sec:spatial_metrics}

\paragraph{Displacements} The distance between two consecutive locations visited by an individual is called jump length or displacement \cite{brockmann2006scaling, gonzalez2008understanding}. 
The term location usually indicates a spatial point in which an individual spent a minimum amount of time reflecting human behavioral tendencies that motivate people to move between two places. 
Formally, a jump length $\Delta r = d(s_i, s_{i+1})$ is the distance between two spatio-temporal points $s_i$ and $s_{i+1}$ in a trajectory $T_u=\langle s_{1}, s_{2}, ..., s_{n} \rangle$. 
A truncated power-law well approximates the empirical distribution $P(\Delta r)$ within a population of individuals, with the 
value of the exponent slightly varying based on the type of data and the spatial scale \cite{brockmann2006scaling, gonzalez2008understanding}. 

\paragraph{Radius of gyration}
The characteristic distance traveled by an individual $u$ during a period of time can be quantified by their radius of gyration \cite{gonzalez2008understanding}, defined as $r_g(u) = \sqrt{\frac{1}{n_u} \sum_{i=1}^{n_u} d(s_i, s_{cm})^2}$, where $n_u$ is the number of points in $T_u$, $s_i \in T_u$ and $s_{cm} = \frac{1}{n_u} \sum_{i=1}^{n_u} s_i$ is the position vector of the center of mass of the set of points in $T_u$.
A truncated power-law well approximates the distribution of $r_g$ \cite{gonzalez2008understanding, pappalardo2013understanding}. 
At a collective level, the evolution in time of the average $r_g$ of individuals follows a logarithmic curve $\langle r_g(t)\rangle \sim \alpha + \beta \ln t$ \cite{gonzalez2008understanding, song2010modelling}.

The $k$-radius of gyration of an individual $u$ is defined as the radius over their $k$ most frequented locations \cite{pappalardo2015returners}, $r_g^{(k)}(u) = \sqrt{\frac{1}{N_k} \sum_{i=1}^k n_i d(s_i - s_{cm}^{(k)})^2}$, where $s_i \in T_u$, $N_k$ is the sum of the visits to $u$'s $k$ most frequented locations, and $s_{cm}^{(k)}  = \frac{1}{N_k} \sum_{i=1}^k s_i$ is the center of mass computed on $u$'s $k$ most frequented locations.
The comparison of $r_g$ and $r_g^{(k)}$ over an entire population revealed the existence of a returners and explorers dichotomy \cite{pappalardo2015returners}.

\paragraph{Mobility entropy}
The temporal-uncorrelated entropy of an individual $u$ characterizes the predictability of their spatial movements, and it is defined as $S_{unc}(u) = - \sum_{i=1}^{n_u} p_u(i) \log_2 p_u(i)$, where $n_u$ is the number of distinct locations visited by $u$ and $p_u(i)$ is the probability that $u$ visits location $i$ \cite{eagle2009eigenbehaviors, song2010limits}.
The real entropy of an individual $u$ considers also the order in which the places were visited and the time spent at each location, and it is defined as $S(u) = - \sum_{T_u' \subset T_u} P(T_u') \log_2 P(T_u')$ \cite{song2010limits}, where $P(T_u')$ is the probability of finding a particular time-ordered subsequence $T_u'$ in the trajectory $T_u$.
The distribution of both $S_{unc}$ and $S$ are peaked and in particular
$P(S)$ peaks around $S \approx 0.8$, indicating that the real spatio-temporal uncertainty in a typical user's whereabouts is $2^{0.8}=1.74$, i.e., fewer than two locations \cite{song2010limits}.

\paragraph{I-rank and G-rank}
Location ranks identify the importance of a location to an individual's mobility: the most visited location (likely home or work) has rank 1, the second most visited location (e.g., school or local shop) has rank 2, and so on. 
The visitation frequency of locations $P(L)$, or I-rank, follows a Zipf law: the probability of finding an individual at a location of rank $L$ is well approximated by $P(L) \sim 1/L$ \cite{gonzalez2008understanding, pappalardo2013understanding}.
Similarly, the collective visitation frequency of a location $P(r)$, or G-rank, indicates the popularity of locations according to how people visit them on the geographic space \cite{pappalardo2017data, ouyang2018non}.

\paragraph{Semantic importance}

For an individual's trajectory $T_u$,
we define $p_{d_{\text{total}}}(r) = \frac{d_{total}(r)}{\sum_{r^{'}} d_{total}(r^{'})}$, 
where $d_{total}(r)$ is the total stay duration in location $r$ interpreted as $r$'s
semantic importance \cite{ouyang2018non, bindeschaedler2016synthesizing}.
The semantic distance between two trajectories $T_u$ and $T_v$ is the distance between the distribution of the $p_{d_{total}}(r)$ for $u$ and $v$.

\paragraph{Mean Distance Error.} Given two equal-sized sets of trajectories $\mathcal{T} = \{T_1, \dots, T_N \}$ and $\mathcal{\hat{T}} = \{ \hat{T}_1, \dots, \hat{T}_N \}$, the Mean Distance Error (MDE) between $\mathcal{T}$ and $\mathcal{\hat{T}}$ is defined as $MDE = \frac{\sum_i^N d(T_i, \hat{T_i})}{N}$, where $d$ is the distance between two points  \cite{huang2019variational}.

\paragraph{Mobility networks} 
In an individual mobility network (IMN), nodes represent locations and directed edges represent an individual's trips between locations \cite{schneider2013unravelling, rinzivillo2014purpose}. 
The vast majority of individuals' trips can be described with a limited number of daily motifs which represent the underlying regularities in daily movements  \cite{schneider2013unravelling}.
Individual trajectories may be aggregated to study the flows of individuals between locations at different spatio-temporal scales.
Flows can typically be described by an Origin-Destination (OD) matrix, or mobility network, which has a specific structure and dynamics \cite{simini2012universal}.

\subsection{Temporal metrics}
\label{sec:temporal_metrics}

\paragraph{Waiting time and circadian rhythm}
The waiting time $\Delta t$ is the elapsed time between two consecutive points in the mobility trajectory of an individual $u$, or equivalently as the time spent in a location: $\Delta t = t_i - t_{i-1}$.
Empirically the distribution of waiting times is well approximated by a truncated power-law \cite{song2010modelling}
The movements of individuals are not distributed uniformly during the hours of the day but follow a circadian rhythm \cite{pappalardo2017data, karamshuk2011human}; people tend to be stationary during the night hours while prefer moving at specific times of the day, for example, to reach the workplace or return home.

\paragraph{Temporal location patterns.}
The temporal popularity $p(r, t)$ measures the visiting probability for a location $r$ at any time $t$ \cite{ouyang2018non}. 
The staying patterns $p(r, d)$ measures the probability of visiting a location $r$ for a duration $d$ \cite{ouyang2018non}.

\end{document}